\documentclass[journal]{IEEEtai}

\usepackage[colorlinks,urlcolor=blue,linkcolor=blue,citecolor=blue]{hyperref}

\usepackage{color,array}

\usepackage{graphicx}

\usepackage{amssymb}
\usepackage[linesnumbered,ruled,vlined]{algorithm2e}
\usepackage{amsmath}
\newcommand{\ie}{\textit{i.e.}, }
\newcommand{\eg}{\textit{e.g.}, }
\usepackage{array}
\usepackage[caption=false,font=normalsize,labelfont=sf,textfont=sf]{subfig}
\usepackage{textcomp}
\usepackage{stfloats}
\usepackage{url}
\usepackage{verbatim}
\usepackage{graphicx}
\usepackage{color,xcolor}
\usepackage{url}
\newtheorem{corollary}{Corollary}
\newtheorem{proof}{Proof}
\usepackage{textcomp}
\usepackage[switch]{lineno}

\usepackage{xcolor}
\usepackage{threeparttable}
\usepackage{booktabs,multicol,mathtools,amsfonts,array,multirow}
\newtheorem{theorem}{Theorem}

\begin{document}

\title{Towards Robust Nonlinear Subspace Clustering: \\ A Kernel Learning Approach}

\author{Kunpeng Xu, Lifei Chen, Shengrui Wang \\
Université de Sherbrooke
}

\markboth{Journal of IEEE Transactions on Artificial Intelligence, Vol. 00, No. 0, Month}
{First A. Author \MakeLowercase{\textit{et al.}}: Bare Demo of IEEEtai.cls for IEEE Journals of IEEE Transactions on Artificial Intelligence}

\maketitle

\begin{abstract}
Kernel-based subspace clustering, which addresses the nonlinear structures in data, is an evolving area of research. Despite noteworthy progressions, prevailing methodologies predominantly grapple with limitations relating to (i) the influence of predefined kernels on model performance; (ii) the difficulty of preserving the original manifold structures in the nonlinear space; (iii) the dependency of spectral-type strategies on the ideal block diagonal structure of the affinity matrix. This paper presents DKLM, a novel paradigm for kernel-induced nonlinear subspace clustering. DKLM provides a data-driven approach that directly learns the kernel from the data's self-representation, ensuring adaptive weighting and satisfying the multiplicative triangle inequality constraint, which enhances the robustness of the learned kernel. By leveraging this learned kernel, DKLM preserves the local manifold structure of data in a nonlinear space while promoting the formation of an optimal block-diagonal affinity matrix. A thorough theoretical examination of DKLM reveals its relationship with existing clustering paradigms. Comprehensive experiments on synthetic and real-world datasets demonstrate the effectiveness of the proposed method.
\end{abstract}

\begin{IEEEImpStatement}
Kernel-based subspace clustering is critical for analyzing nonlinear data structures, yet existing methods face challenges with predefined kernels, loss of manifold structure, and inadequate block-diagonal affinity matrices. These limitations reduce their effectiveness in applications such as motion segmentation and image clustering. In this paper, we introduce DKLM, a data-driven kernel subspace clustering method that learns the kernel directly from data. Unlike traditional approaches, DKLM preserves the local manifold structure and promotes a robust block-diagonal affinity matrix, enhancing clustering accuracy. Experimental results show that DKLM outperforms SOTA methods across synthetic and real-world datasets, particularly in scenarios with complex nonlinear patterns, such as face images with varying illumination and motion trajectories with perspective distortion. DKLM also proves effective for high-dimensional time series data, paving the way for its application in domains like computer vision, bioinformatics, and financial analytics, where robust handling of nonlinear data is essential.
\end{IEEEImpStatement}

\begin{IEEEkeywords}
Subspace clustering, Data-driven kernel learning, Local manifold structure, Nonlinear self-representation
\end{IEEEkeywords}

\section{Introduction}
High-dimensional data clustering plays a crucial role in uncovering patterns across various domains of data mining and machine learning. Typically, high-dimensional data tends to be distributed across several low-dimensional subspaces. For instance, facial images under diverse lighting circumstances are often represented by a 9-dimensional linear subspace. Numerous subspace clustering techniques have been established to identify these subspaces, encompassing statistical analysis \cite{rao2008motion}, algebraic operators \cite{agarwal2004k}, iterative optimization \cite{vidal2005generalized}, deep learning \cite{ji2017deep}, and spectral methods \cite{lu2018subspace}. The process of subspace clustering, which involves partitioning data points from a union of linear subspaces into their respective subspaces, has been applied in diverse fields, \eg motion segmentation \cite{costeira1998multibody,keuper2018motion}, gene expression profiling in bioinformatics \cite{xu2022multi}, identifying financial market regimes \cite{lim2019temporal}, and detecting social network communities  \cite{jia2020constrained,yang2019subspace}.



Among these current methodologies, spectral-type approaches have gained prominence because of their solid mathematical foundation and effectiveness. Such methods typically consist of two steps \cite{lu2018subspace,zhu2018low}: (i) creating an affinity matrix from the self representation of the data, capturing the similarities among data points, and (ii) applying spectral clustering to partition data into their respective subspaces. However, real-world data often deviates from linear subspace models. For face image clustering, reflectance frequently exhibits non-Lambertian characteristics, and the position of subjects may fluctuate, while in motion segmentation, perspective distortion in the camera can invalidate the affine camera assumption. Consequently, both face images and motion trajectories generally reside in a nonlinear subspace.


Recent research trends have explored kernel learning as a promising alternative \cite{yin2016kernel,ji2017adaptive,xue2020robust,xu2022data}. For example, the study in \cite{patel2014kernel} adapted sparse subspace clustering (SSC) \cite{elhamifar2013sparse} via substituting the inner product of the data matrix with polynomial or RBF kernels. Similarly, \cite{nguyen2015kernel,xiao2015robust} employed kernelization in the low-rank representation (LRR) method \cite{liu2012robust}, whereas \cite{yin2016kernel} applied the Log-Euclidean kernel to symmetric positive definite matrices to enhance SSC. Nonetheless, kernel selection is predominantly empirical, and there is insufficient data indicating that the implicit feature space of a predetermined kernel is necessarily appropriate for subspace clustering. Although multiple kernel learning (MKL) attempts to overcome the limitation by integrating different kernel functions, it still relies on the assumption that data originates from a mixture of these predefined distributions. Furthermore, these approaches often neglect preserving the local manifold structures \footnote{In this paper, our scope extends beyond data structures nonlinearly deviate from linear subspaces, encompassing clusters that can lie arbitrarily far from linear subspaces, such as arbitrary manifold (e.g., Olympic rings and spirals).} and connectivity within each nonlinear subspace, thereby affecting the effectiveness of subsequent spectral clustering -- particularly since most real-world data do not satisfy the independent subspace assumption.



To overcome these limitations, we present DKLM, a data-driven model for kernel subspace clustering that ensures the preservation of local manifold structures. Instead of relying on predefined kernel functions, our approach learns the kernel directly from the data’s self-representation, circumventing this fundamental obstacle. We derive self-representation with specific conditions that can be viewed as an adaptive-weighting kernel learning while satisfying the triangular inequality constraint. Additionally, the learned kernel facilitates the discovery of improved subspaces with a block-diagonal structure. Our approach is general and could be implemented within various self-representation-based subspace clustering frameworks. Theoretical analysis demonstrates DKLM’s connection to existing methods, and experiments show that DKLM achieves superior clustering performance compared to state-of-the-art approaches. 

The main contributions of this work are as follows:
\begin{itemize}
    \item[1)] We present a data-driven kernel subspace clustering method that adaptively learns a kernel mapping without predefined functions, while maintaining both manifold structures and block diagonal properties.
    \item[2)]  We redefine kernel learning as a self representation learning problem, adhering to adaptive weighting and the multiplicative triangular inequality constraint. We provide the theoretical analysis of DKLM and the connections to other clustering algorithms. To the best of our knowledge, this work is the first to derive the kernel directly from the self-representation of the data.
    \item[3)] DKLM consistently outperforms state-of-the-art self representation and kernel-based subspace clustering methods by producing sparse, symmetric, nonnegative, and block-diagonal solutions. Extensive experiments on four synthetic and nine real-world datasets confirm the accuracy and robustness of our approach.
\end{itemize}
We also extend our experiments to several high-dimensional time series data in different applications and demonstrate the practicality and effectiveness of our method for discovering nonlinear patterns in time series (Section~\ref{sec:ts}).

\noindent \textbf{Outline.} The remainder of this paper is structured as follows: we begin with a review of related work, followed by our proposed model, experimental results, and finally, the conclusions.

\section{Related Work}
This section provides an overview of mainstream methods that are most relevant to our work. While a comprehensive survey lies beyond the scope of this paper, we offer a critical review to better position our approach within the broader research landscape. A set of commonly used parameters is summarized in Table~\ref{tab:notations}.
\begin{table}[h!]
\centering
\caption{Commonly used notations and symbols.}
\label{tab:notations}
\resizebox{0.48\textwidth}{!}{%
\begin{tabular}{ll|ll}
\toprule
\textbf{Notation} & \textbf{Definition} & \textbf{Notation} & \textbf{Definition} \\
\midrule
\multicolumn{4}{l}{\textit{General Matrices and Vectors}} \\
\midrule
$\mathbf{M}$ & Matrix & $\mathbf{m}$ & Vector \\
$\mathbf{M}_{ij}$ & $(i,j)$-th entry of $\mathbf{M}$ & $\mathbf{M}^\mathrm{T}$ & Transpose of $\mathbf{M}$ \\
$\mathbf{M}^{-1}$ & Inverse of $\mathbf{M}$ & $\mathrm{Tr}(\mathbf{M})$ & Trace of $\mathbf{M}$ \\
$\mathrm{diag}(\mathbf{M})$ & Diagonal elements of $\mathbf{M}$ & $\mathrm{Diag}(\mathbf{m})$ & Diagonal matrix from $\mathbf{m}$ \\
$\mathbf{I}$ & Identity matrix & $\mathbf{1}$ & All-ones vector \\
$[\mathbf{M}]_+$ & Nonnegative part of $\mathbf{M}$ & $\mathbf{M} \succeq 0$ & Positive semi-definite matrix \\
\midrule
\multicolumn{4}{l}{\textit{Problem-Specific Matrices and Parameters}} \\
\midrule
$N$ & Number of data points &$D$ & Feature dimension \\
$Q$ & Number of random points & $\alpha$ & Local structure weight \\
$\gamma$ & Regularization weight & $\beta$ & Relaxation weight \\
$\mathbf{X}$ & Data matrix $(\mathcal{R}^{D \times N})$ & $\widehat{\mathbf{X}}$ & Random selection set \\
$\boldsymbol{\mathcal{K}}$ & Kernel matrix & $\boldsymbol{\mathcal{K}}(\cdot, \cdot)$ & Kernel measure \\
$\boldsymbol{\widetilde{\mathcal{K}}}$ & Approximated kernel matrix & $\Phi(\cdot)$ & Mapping function \\
$\mathbf{Z}$ & Self-representation matrix & $\mathbf{C}$ & Relaxation matrix for $\mathbf{Z}$ \\
$\mathbf{S}$ & Auxiliary matrix for $\mathbf{C}$ & $\mathbf{W}$ & Affinity matrix $(\mathbf{Z}^\mathrm{T} + \mathbf{Z}) / 2$ \\
$\mathbf{G}$ & Degree-based matrix & $\mathbf{L_Z}$ & Laplacian matrix of $\mathbf{Z}$ \\
$\mathbf{P}$ & Permutation matrix & $\|\cdot\|$ & Frobenius norm \\
$\|\cdot\|_{\boxed{\scriptstyle k}}$ & $k$-block diagonal norm & $\mathrm{Reg}(\cdot)$ & Regularization term \\
\bottomrule
\end{tabular}}
\end{table}

\subsection{Linear Subspace Clustering}
Traditional subspace clustering methods emphasize the construction of an affinity matrix where data points belonging to the same subspace exhibit strong similarity, while points from distinct subspaces should indicate minimum or no affinity. Recent approaches based on self-representation \cite{elhamifar2013sparse,lu2018subspace,bai2020sparse} leverage the principle of subspace self-representation, in which each point expresses a point within a subspace as a linear combination of other points from the same subspace, employing the resultant self-representation matrix as the affinity matrix for spectral clustering.
Specifically, for a data matrix $\mathbf{X} \in \mathcal{R}^{D \times N}$, where each column corresponds to a single data point, $\mathbf{X}$ can be decomposed as $\mathbf{X = XZ}$, with $\mathbf{Z}$ as the self-representation coefficient matrix. The subspace clustering problem can be formulated as the following optimization task:
\begin{equation}
    \min_{\mathbf{Z}} \frac{1}{2}||\mathbf{X-XZ}||^2+\gamma \mathrm{Reg}(\mathbf{Z}),\ s.t. \ \mathbf{Z}\geq 0, \mathrm{diag}(\mathbf{Z})=0 
\label{eq:subclustering}
\end{equation}
Here, $\mathrm{Reg}(\mathbf{Z})$ represents a regularization term for $\mathbf{Z}$, $\gamma > 0$ controls the balance between the reconstruction error and regularization, and the constraint $\mathrm{diag}(\mathbf{Z}) = 0$ ensures that the trivial identity solution is avoided. Additionally, the constraint $\mathbf{1}^\mathrm{T}\mathbf{Z} = \mathbf{1}^\mathrm{T}$ is included to ensure that the data lies within a union of subspaces. Based on this self-representation framework, the spectral clustering affinity matrix can be derived as $\mathbf{W = (Z}^\mathrm{T} + \mathbf{Z}) / 2$. The primary distinction among existing methods comes from the selection of $\mathrm{Reg}(\mathbf{Z})$, which can promote sparsity (SSC \cite{elhamifar2013sparse}, $||\cdot||_1$), least-squares fitting (LSR \cite{lu2012robust}, $||\cdot||^2$), low-rankness (LRR \cite{liu2012robust}, $||\cdot||_*$), or entropy constraints (SSCE \cite{bai2020sparse}, $||\cdot||_{E}$). Ideally, for $N$ points distributed across $k$ clusters, $\mathbf{Z}$ should satisfy two critical properties: First, the affinities between different clusters should be zero, ensuring that the within-cluster affinities accurately reflect the similarity between samples. This structure allows $\mathbf{Z}$ to represent the global organization of the data \cite{bai2020sparse}. Second, $\mathbf{Z}$ should exhibit a block diagonal structure, with exactly $k$ interconnected blocks. Each block corresponds to a unique subspace, and these blocks should be visually dense and distinct \cite{ren2020simultaneous}. 
\subsection{Nonlinear Subspace Clustering}

Several studies have extended subspace clustering methods from linear to nonlinear contexts through the ``kernel trick". For example, in order to regularize the representation error, \cite{patel2014kernel} and \cite{wang2011structural} applied the Frobenius norm to SSC and LRR, respectively, in the kernel space. Xiao et al. \cite{xiao2015robust} introduced a robust kernelized variant of LRR with a closed-form solution for the subproblem. Ji et al. \cite{ji2017adaptive} presented a kernel approximation technique that learns a predefined kernel function. Kang et al. \cite{kang2017twin} presented a framework for jointly learning the cluster indicator matrix and similarity metrics, while Yang et al. \cite{yang2019joint} developed a joint correntropy metric with block diagonal regularization for robust kernel clustering.

Multiple Kernel Learning (MKL) has been extensively utilized to overcome the limitations of relying on a single predefined kernel. By combining multiple kernels, MKL can better capture the underlying structure of the data. For example, Ren and Sun \cite{ren2020simultaneous} proposed an MKL method that preserves both global and local graph structures, while Liu et al. \cite{liu2024local} developed a localized kernel-based graph learning approach for MKL. Tang et al. \cite{tang2023knowledge} introduced a knowledge-induced fuzzy clustering approach that integrates domain knowledge into MKL. Liu \cite{liu2023hyperparameter} proposed a hyperparameter-free MKL method that ensures global optimality in localized kernel $k$-means clustering. Additionally, Su et al. \cite{su2024kernel} presented a kernel correlation-dissimilarity measure for MKL k-means clustering.

However, despite these advancements, MKL methods face challenges. Their effectiveness heavily depends on the initial kernel function selection, which is often empirical and may not align with the actual data distribution. The combined distribution of these kernels may fail to fully capture data complexity, leading to suboptimal performance. Additionally, MKL methods generally neglect the preservation of local manifold structures in the data, potentially compromising clustering results, especially in complex nonlinear structures.


\section{Proposed Method}
This section presents a novel data-driven kernel subspace clustering model that maintains the local manifold structure while projecting data onto a high-dimensional reproducing kernel Hilbert space (RKHS).
\subsection{Kernel Subspace Clustering with Local Manifold Preservation}
We map the data into the RKHS in order to handle the nonlinear structure, which makes linear pattern analysis more feasible. By jointly leveraging the advantages of kernel mapping and self-representation, we reformulate the optimization problem \eqref{eq:subclustering} as follows:
\begin{equation}
\begin{aligned}
    \min_{\mathbf{Z}} &\frac{1}{2}|| \mathrm{\Phi}(\mathbf{X})-\mathrm{\Phi}(\mathbf{X})\mathbf{Z}||^2+\gamma\text{Reg}(\mathbf{Z}) \\ & =\min_{\mathbf{Z}} \frac{1}{2}\mathrm{Tr}(\mathrm{\Phi}(\mathbf{X})^\mathrm{T}\mathrm{\Phi}(\mathbf{X})-2\mathrm{\Phi}(\mathbf{X})^\mathrm{T}\mathrm{\Phi}(\mathbf{X})\mathbf{Z}\\& \quad +\mathbf{Z}^\mathrm{T}\mathrm{\Phi}(\mathbf{X})^\mathrm{T}\mathrm{\Phi}(\mathbf{X})\mathbf{Z})+ \gamma\text{Reg}(\mathbf{Z}) \\&=\min_{\mathbf{Z}} \frac{1}{2}\mathrm{Tr}(\boldsymbol{\mathcal{K}}-2\boldsymbol{\mathcal{K}}\mathbf{Z}+\mathbf{Z}^\mathrm{T}\boldsymbol{\mathcal{K}}\mathbf{Z})+\gamma\text{Reg}(\mathbf{Z}),
    \\ & \quad \quad \quad s.t.\ \mathbf{Z} \geq 0, \mathrm{diag}(\mathbf{Z})=0, \mathbf{{1}^{\mathrm{T}}\mathbf{Z}=\mathbf{1}^{\mathrm{T}}}
\end{aligned}
\label{eq:KernelSub}
\end{equation}
where the mapping function $\mathrm{\Phi}(\cdot)$ is implicitly handled by using a kernel matrix $\boldsymbol{\mathcal{K}}$, which satisfies  $\boldsymbol{\mathcal{K}}=\mathrm{\Phi}(\cdot)^{\mathrm{T}}\mathrm{\Phi}(\cdot)$, defining the similarity metric and adhering to the following properties:

\noindent\textbf{Condition 1 (Non-negativity):} $\boldsymbol{\mathcal{K}}(\mathbf{X}_i, \mathbf{X}_j) \geq 0$,

\noindent\textbf{Condition 2 (Symmetry):} $\boldsymbol{\mathcal{K}}(\mathbf{X}_i, \mathbf{X}_j) = \boldsymbol{\mathcal{K}}(\mathbf{X}_j, \mathbf{X}_i)$,

\noindent\textbf{Condition 3 (Positive semi-definite):} $\boldsymbol{\mathcal{K}} \succeq 0$.

The objective is to create an affinity matrix with precisely $k$ connected components if $\mathbf{X}$ comprises $k$ clusters. This indicates that, with suitable permutations, the self-representation matrix $\mathbf{Z}$ would display a $k$-block diagonal structure. The representation matrix $\mathbf{Z}$ derived from \eqref{eq:KernelSub} may not inherently exhibit this property. Fortunately, \textit{Theorem~\ref{theo:bdr}} guarantees the emergence of $k$ connected components by linking the eigenvalues of the Laplacian matrix to the clustering structure.

\begin{theorem}[\cite{von2007tutorial}, Proposition 4]\label{theo:bdr}
\textit{The number of connected components (blocks) in $\mathbf{Z}$} is equal to the multiplicity $k$ of the eigenvalue 0 of the associated Laplacian matrix $\mathbf{L_Z}$ for any $\mathbf{Z}\ge0, \mathbf{Z}=\mathbf{Z}^T$.
\end{theorem}

In accordance with this theorem and following the work \cite{lu2018subspace}, we introduce a block diagonal regularizer (BDR) in nonlinear space to encourage $\mathbf{Z}$ to achieve a block diagonal structure when the underlying subspaces are independent. Incorporating the BDR, our model is reformulated as:
\begin{equation}
\begin{aligned}
    \min_{\mathbf{Z}} &\frac{1}{2}\mathrm{Tr}(\boldsymbol{\mathcal{K}}-2\boldsymbol{\mathcal{K}}\mathbf{Z}+\mathbf{Z}^\mathrm{T}\boldsymbol{\mathcal{K}}\mathbf{Z})+\gamma ||\mathbf{Z}||_{\boxed{\scriptstyle k}}, \\&s.t.\ \mathbf{Z} = \mathbf{Z}^\mathrm{T} \geq 0, \mathrm{diag}(\mathbf{Z})=0, \mathbf{{1}^{\mathrm{T}}\mathbf{Z}=\mathbf{1}^{\mathrm{T}}}
\end{aligned}
\label{eq:kbdr}
\end{equation}
where
\begin{equation}
    ||\mathbf{Z}||_{\boxed{\scriptstyle k}} = \sum_{i=N-k+1}^N \lambda_i(\mathbf{L_{\mathbf{Z}}})
\nonumber
\end{equation}
The eigenvalues of $\mathbf{L_Z}$, denoted as $\lambda_i(\mathbf{L_Z})$, are arranged in descending order. Notably, the condition $||\mathbf{Z}||_{\boxed{\scriptstyle k}}=0$ implies that $\mathbf{Z}$ possesses a $k$-block diagonal structure.

Additionally, the negative term $-\mathrm{Tr}(\boldsymbol{\mathcal{K}}\mathbf{Z})$ in \eqref{eq:kbdr} can be reformulated as:
\begin{equation}
\begin{aligned}
    \min_{\mathbf{Z}} -\mathrm{Tr}(\boldsymbol{\mathcal{K}}\mathbf{Z})&=\min_{\mathbf{Z}}\sum_{i=1}^N \sum_{j=1}^N -\mathrm{\Phi}(\mathbf{X}_i)^\mathrm{T}\mathrm{\Phi}(\mathbf{X}_j)\mathbf{Z}_{ij}\\ &=\min_{\mathbf{Z}}\sum_{i=1}^N \sum_{j=1}^N -\boldsymbol{\mathcal{K}}(\mathbf{X}_i,\mathbf{X}_j)\mathbf{Z}_{ij},
    \\& s.t.\ \mathbf{Z} = \mathbf{Z}^\mathrm{T} \geq 0, \mathrm{diag}(\mathbf{Z})=0, \mathbf{{1}^{\mathrm{T}}\mathbf{Z}=\mathbf{1}^{\mathrm{T}}} 
\end{aligned}
\label{eq:kz}
\end{equation}
in which the similarity between $\mathbf{X}_i$ and $\mathbf{X}_j$ in kernel space is represented by $\boldsymbol{\mathcal{K}}(\mathbf{X}_i,\mathbf{X}_j)$. From \eqref{eq:kz}, it is apparent that increased similarity (i.e., reduced distance) $\boldsymbol{\mathcal{K}}(\mathbf{X}_i,\mathbf{X}_j)$ typically results in a greater $\mathbf{Z}_{ij}$, and conversely. This acts as a kernel extension for maintaining the local manifold structure, akin to minimizing $\min_{\mathbf{Z}}\sum_{i=1}^N\sum_{j=1}^N||\mathbf{X}_i-\mathbf{X}_j||^2\mathbf{Z}_{ij}$ in linear space \cite{nie2014clustering,zhan2018graph}. By allocating a diminished weight to this negative component, the self-representation process will integrate contributions from all data points. In contrast, it will emphasize contributions from neighbors, enhancing the sparsity of $\mathbf{Z}$ while preserving the local manifold structure. The impact of the term $-\mathrm{Tr}(\boldsymbol{\mathcal{K}}\mathbf{Z})$ is best fine-tuned for optimal results. The ultimate objective function is achieved by integrating kernel mapping, BDR, and local manifold preservation into a cohesive framework:
\begin{equation}
\begin{aligned}
     \min_{\mathbf{Z}} &\frac{1}{2}\mathrm{Tr}(\boldsymbol{\mathcal{K}}+\mathbf{Z}^\mathrm{T}\boldsymbol{\mathcal{K}}\mathbf{Z})-\alpha\mathrm{Tr}(\boldsymbol{\mathcal{K}}\mathbf{Z})+\gamma ||\mathbf{Z}||_{\boxed{\scriptstyle k}}, \\&s.t.\ \mathbf{Z} = \mathbf{Z}^\mathrm{T} \geq 0, \mathrm{diag}(\mathbf{Z})=0, \mathbf{{1}^{\mathrm{T}}\mathbf{Z}=\mathbf{1}^{\mathrm{T}}}
\end{aligned}
\label{eq:obj}
\end{equation}
Here $\alpha, \gamma > 0 $ balance local structure preservation and block diagonal properties, respectively.

\noindent \textbf{Optimization}
To define the block diagonal regularizer, the matrix $\mathbf{Z}$ is required to be both nonnegative and symmetric in the formulation \eqref{eq:obj}. These constraints, nevertheless, might limit its expressive potential. To address this, we introduce an auxiliary matrix $\mathbf{C}$, reformulating \eqref{eq:obj} as:
\begin{equation}
\begin{split}
\min_{\mathbf{Z,C}} \frac{1}{2}\mathrm{Tr}(\boldsymbol{\mathcal{K}}+&\mathbf{Z}^\mathrm{T}\boldsymbol{\mathcal{K}}\mathbf{Z})-\alpha\mathrm{Tr}(\boldsymbol{\mathcal{K}}\mathbf{Z}) +\frac{\beta}{2}||\mathbf{Z-C}||^2+\gamma ||\mathbf{C}||_{\boxed{\scriptstyle k}}, \\& s.t.\ \mathbf{C} = \mathbf{C}^\mathrm{T} \geq 0, \mathrm{diag}(\mathbf{C})=0, \mathbf{{1}^{\mathrm{T}}\mathbf{C}=\mathbf{1}^{\mathrm{T}}}
\end{split}
\label{eq:objrelax}
\end{equation}
When $\beta > 0$ is sufficiently large, the two models mentioned above, \eqref{eq:obj} and \eqref{eq:objrelax}, become equivalent. Moreover, the objective function is separable due to the relaxation term $||\mathbf{Z} - \mathbf{C}||^2$. This further guarantees that there will be stable and unique solutions since the subproblems for updating $\mathbf{Z}$ and $\mathbf{C}$ remain strongly convex.

The term $||\mathbf{C}||_{\boxed{\scriptstyle k}}$ is nonconvex, and following \cite{dattorro2010convex}, we have $\sum_{i=N-k+1}^N \lambda_i(\mathbf{L}) = \min_{\mathbf{S}} <\mathbf{L}, \mathbf{S}>$, where $0 \preceq \mathbf{S} \preceq \mathbf{I}$ and $\mathrm{Tr}(\mathbf{S}) = k$. Thus, \eqref{eq:objrelax} becomes:
\begin{equation}
\begin{split}
&\min_{\mathbf{Z},\mathbf{C},\mathbf{S}} \frac{1}{2}\mathrm{Tr}(\boldsymbol{\mathcal{K}}+\mathbf{Z}^\mathrm{T}\boldsymbol{\mathcal{K}}\mathbf{Z})-\alpha\mathrm{Tr}(\boldsymbol{\mathcal{K}}\mathbf{Z}) +\frac{\beta}{2}||\mathbf{Z-C}||^2\\&\quad \quad \quad \quad \quad \quad \quad+\gamma<\mathrm{Diag}(\mathbf{C1})-\mathbf{C},\mathbf{S}> \\  & \quad \quad \quad s.t.\ \mathbf{C} = \mathbf{C}^\mathrm{T} \geq 0, \mathrm{diag}(\mathbf{C})=0, \mathbf{{1}^{\mathrm{T}}\mathbf{C}=\mathbf{1}^{\mathrm{T}}},\\& \quad\quad\quad\quad\quad0 \preceq \mathbf{S} \preceq \mathbf{I}, \mathrm{Tr}(\mathbf{S})=\textit{k}
\end{split}
\label{eq:zcw}
\end{equation}
Equation (\eqref{eq:zcw}) is convex for each variable while the others are held fixed, even though it is not jointly convex for $\mathbf{Z}$, $\mathbf{S}$, and $\mathbf{C}$. Therefore, it can be efficiently solved using alternating optimization steps, as outlined below.

\noindent\textbf{First, the optimization of} $\mathbf{Z}$ is performed by keeping $\mathbf{C} = \mathbf{C}^{i}$ and $\mathbf{S} = \mathbf{S}^{i}$ fixed. The goal is to minimize the following objective function:

\begin{equation}
\mathbf{Z}^{i+1} = \arg \min_{\mathbf{Z}} \frac{1}{2}\mathrm{Tr}(\boldsymbol{\mathcal{K}} + \mathbf{Z}^\mathrm{T}\boldsymbol{\mathcal{K}} \mathbf{Z}) - \alpha\mathrm{Tr}(\boldsymbol{\mathcal{K}} \mathbf{Z}) + \frac{\beta}{2} ||\mathbf{Z} - \mathbf{C}||^2
\label{eq:upZ}
\end{equation}

This leads to the solution:

\begin{equation}
\begin{split}
\mathbf{Z}^{i+1} &=(\frac{1}{2}\boldsymbol{\mathcal{K}}+\frac{1}{2}\boldsymbol{\mathcal{K}}^{\mathrm{T}}+\beta \mathbf{I})^{\mathrm{-1}}(\alpha \boldsymbol{\mathcal{K}}^{\mathrm{T}}+\beta \mathbf{C}) \\ &= (\boldsymbol{\mathcal{K}} + \beta \mathbf{I})^{-1}(\alpha \boldsymbol{\mathcal{K}} + \beta \mathbf{C})
\end{split}
\end{equation}

\noindent \textbf{Next, with $\mathbf{Z}$ updated, the focus shifts to} $\mathbf{S}$. By fixing $\mathbf{C} = \mathbf{C}^{i}$, $\mathbf{S}$ is updated by solving:

\begin{equation}
\begin{split}
\mathbf{S}^{i+1} = \arg \min_{\mathbf{S}} \left< \mathrm{Diag}(\mathbf{C1}) - \mathbf{C}, \mathbf{S} \right>
\end{split}
\label{eq:upW}
\end{equation}

while adhering to the constraints $0 \preceq \mathbf{S} \preceq \mathbf{I}$ and $\mathrm{Tr}(\mathbf{S}) = k$. The solution involves eigen-decomposition, resulting in $
\mathbf{S}^{i+1} = \mathbf{U} \mathbf{U}^{\mathrm{T}}$, where $\mathbf{U}$ is formed from the $k$ eigenvectors corresponding to the smallest eigenvalues of $\mathrm{Diag}(\mathbf{C1}) - \mathbf{C}$.

\noindent\textbf{Finally, to update} $\mathbf{C}$, with $\mathbf{S} = \mathbf{S}^{i+1}$ and $\mathbf{Z} = \mathbf{Z}^{i+1}$ fixed, the objective is to minimize:

\begin{equation}
\begin{aligned}
\mathbf{C}^{i+1} = \arg \min_{\mathbf{C}} \frac{\beta}{2}||\mathbf{Z} - \mathbf{C}||^2 + \gamma \left< \mathrm{Diag}(\mathbf{C1}) - \mathbf{C}, \mathbf{S} \right>
\end{aligned}
\end{equation}

which simplifies to:

\begin{equation}
\begin{aligned}
\mathbf{C}^{i+1} = \arg \min_{\mathbf{C}} \frac{1}{2}||\mathbf{C} - \mathbf{Z} + \frac{\gamma}{\beta}(\mathrm{diag}(\mathbf{S}) \mathbf{1}^{\mathrm{T}} - \mathbf{S})||^2
\end{aligned}
\label{eq:upC}
\end{equation}

The closed-form solution for this step is:

\[
\mathbf{C}^{i+1} = [(\mathbf{\hat{A}} + \mathbf{\hat{A}}^{\mathrm{T}})]_{+}
\]

where $\mathbf{\hat{A}} = \mathbf{A} - \mathrm{Diag}(\mathrm{diag}(\mathbf{A}))$ and $\mathbf{A} = \mathbf{Z} - \frac{\gamma}{\beta}(\mathrm{diag}(\mathbf{S}) \mathbf{1}^{\mathrm{T}} - \mathbf{S})$.

The complete algorithm for solving \eqref{eq:zcw} is outlined in Algorithm~\ref{alg}.
\begin{algorithm}[htbp]
\caption{Alternating Minimization to Solve \eqref{eq:zcw}}
\label{alg}
\LinesNumbered
{\small{
\KwIn{ $\boldsymbol{\mathcal{K}}$, trade-off parameters $\alpha$, $\beta$, $\gamma$, number of clusters $k$;
}
\KwOut{Solution matrix $\mathbf{Z}$}
\tcc{Initialization:}
Set iteration index $i=1$, and initialize $\mathbf{S}^i=0$, $\mathbf{Z}^i=0$, $\mathbf{C}^i=0$ \\
\While{convergence is not reached}{
Compute $\mathbf{Z}^{i+1}$ using the closed-form solution from \eqref{eq:upZ};\\
Update $\mathbf{S}^{i+1}$ using the eigenvalue-based closed-form solution from \eqref{eq:upW};\\
Update $\mathbf{C}^{i+1}$ by solving \eqref{eq:upC} in closed form;\\
Increment $i=i+1$.
}
}}
\end{algorithm}

\subsection{Adaptive-weighting Kernel Learning} \label{kernel}
Using $\boldsymbol{\mathcal{K}}$ as input, we obtain the self-representation matrix $\mathbf{Z}$ in the nonlinear space by solving the optimization problem in \eqref{eq:obj}. The selection of a kernel, such as linear, polynomial, or radial basis function (RBF), often has a significant impact on the performance of a kernel-based methods. Instead of predefining a kernel function, we learn the kernel from the self-representation matrix itself. By interpreting the kernel as a similarity metric, we define:
\begin{equation}
    \boldsymbol{\mathcal{K}}_{ij} = \begin{cases} 
\exp(-2 \cdot \max(\mathbf{G}) + \mathbf{G}_{ij}), & \text{if } i \neq j,\\
\sum_{q \neq i} \exp(-2 \cdot \max(\mathbf{G}) + \mathbf{G}_{iq}) + \xi, & \text{if } i = j,
\end{cases}
\label{eq:kernel}
\end{equation}
where 
\begin{equation}
    \mathbf{G} = [\mathrm{Diag}(\hat{\mathbf{W}}\mathbf{1})]^{-\frac{1}{2}}\hat{\mathbf{W}}[\mathrm{Diag}(\hat{\mathbf{W}}\mathbf{1})]^{-\frac{1}{2}}
\label{weight}
\end{equation}

\noindent \textbf{Remark.} 
The semi-positive definiteness of $\boldsymbol{\mathcal{K}}$ is guaranteed by the tuning parameter $\xi \in (0,1)$. The use of  $max(\mathbf{G})$ in non-diagonal elements introduces an adaptive scaling, reflective of the self-representation-derived data structure. This choice facilitates dynamic similarity scaling linked to the properties of the self-representation matrix. By applying $\max(\mathbf{G})$ within the exponential function for dynamic scaling, the similarity measure is adeptly adjusted according to the actual distribution of the data, ensuring accurate mapping of relative positions and similarities between data points, regardless of the data density. Additionally, $\hat{\mathbf{W}}=(\hat{\mathbf{Z}}^{\mathrm{T}}+\hat{\mathbf{Z}})/2$ is derived from solving \eqref{eq:subclustering} with any regularizer in linear space, captures the underlying data structure for constructing a kernel. Equation~\eqref{weight} applies symmetric normalization to $\hat{\mathbf{W}}$, resulting in a symmetric kernel matrix $\boldsymbol{\mathcal{K}}$, which contrasts with traditional asymmetric normalization methods. This symmetric approach also dynamically adjusts and weights the similarity across data points (refer to \textit{Corollary~\ref{coro:degree}}). As stated in \textit{Theorem~\ref{theo:kernel}}, $\xi$ in \eqref{eq:kernel} guarantees that $\boldsymbol{\mathcal{K}}$ satisfies the properties of a kernel matrix.

\begin{figure}[t]
\centerline{\includegraphics[width=1\linewidth]{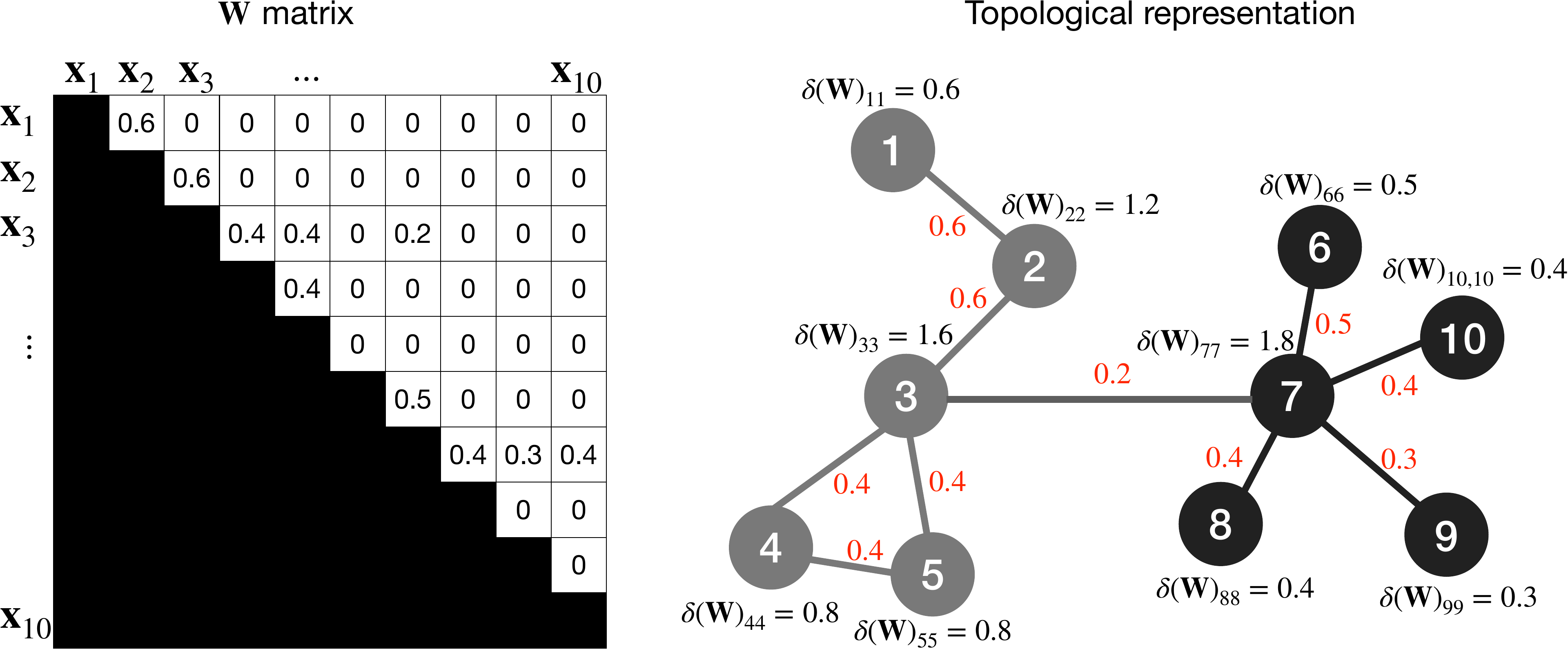}}
\caption{To illustrate Corollary 1, consider data points ($\mathbf{x}_1,\cdots, \mathbf{x}_{10}$), organized into two distinct clusters. Since $\mathbf{W}$ is symmetric, we only need to examine one triangle (upper or lower) for this representation. In the topological view, integers 1–10 correspond to $\mathbf{x}_1\ -\ \mathbf{x}_{10}$, respectively, with edge weights derived from $\mathbf{W}$. Observing the example, nodes 3 and 7, with higher degrees, result in $\mathbf{G}_{37} =0.2/\sqrt{1.6*1.8}\thickapprox 0.1179$, indicating reduced similarity between clusters. Conversely, similarity within the same cluster is enhanced, as shown by $\mathbf{G}_{12}=0.6/\sqrt{0.6*1.2}\thickapprox 0.7071$.}
\label{Fig:graph}
\end{figure}

\begin{corollary}[Degree-scaled weighting] \label{coro:degree}

Since $\mathbf{G} =\delta(\mathbf{W})^{-\frac{1}{2}} \mathbf{W}\delta(\mathbf{W})^{-\frac{1}{2}}$, where $\delta(\mathbf{W}) = \mathrm{Diag}(\mathbf{W} \mathbf{1})$ represents the row-wise degree of $\mathbf{W}$. Then:
\begin{equation}
\begin{aligned}
\mathbf{G}_{ij}&=\delta(\mathbf{W})_{ii}^{-\frac{1}{2}}\mathbf{W}_{ij}\delta(\mathbf{W})_{jj}^{-\frac{1}{2}}=\frac{\mathbf{W}_{ij}}{\sqrt{\delta(\mathbf{W})_{ii}\delta(\mathbf{W})_{jj}}}
\end{aligned}
\nonumber
\end{equation}
The degrees of points $i$ and $j$, denoted as $\delta(\mathbf{W})_{ii}$ and $\delta(\mathbf{W})_{jj}$, have an inverse relationship with the interaction weight between each point. Data points with higher degrees tend to exhibit smaller interaction weights. Typically, weights within the same cluster are larger compared to those between clusters. (\textit{N.b.}: In practice, even data points contributing to the same cluster may occasionally be assigned small weights despite their high similarity; nevertheless, this degree-based weighting effectively reduces within-cluster dispersion). Figure~\ref{Fig:graph} provides a visual representation from a topological perspective.
\end{corollary}  
\begin{theorem}\label{theo:kernel}
The acquired $\rm\boldsymbol{\mathcal{K}}$ is a valid kernel, \ie it meets \textbf{Condition 1 - Condition 3}.
\end{theorem}

\begin{proof}
It is evident that $\boldsymbol{\mathcal{K}}$ satisfies the first two conditions, so we focus on proving its positive semi-definiteness. Since $\boldsymbol{\mathcal{K}}_{ii} - \sum_{j \neq i}|\boldsymbol{\mathcal{K}}_{ij}| = \xi > 0$, we conclude that $\boldsymbol{\mathcal{K}}$ is strictly diagonally dominant, implying it is invertible and symmetric. Let $\mathbf{D} = \mathrm{Diag}(\boldsymbol{\mathcal{K}}_{11}, \cdots, \boldsymbol{\mathcal{K}}_{NN})$ and define a path $\mathbf{M}(t) = (1 - t)(\mathbf{D} + \mathbf{I}) + t\boldsymbol{\mathcal{K}}$ for $t \in [0,1]$. Note that $\mathbf{M}(t)$ remains strictly diagonally dominant for $t < 1$:
\begin{equation}
\begin{aligned}
\sum _{j\neq i}t|\boldsymbol{\mathcal{K}}_{ij}|< t\boldsymbol{\mathcal{K}}_{ii}&<\boldsymbol{\mathcal{K}}_{ii}+(1-t) =(1-t)(\boldsymbol{\mathcal{K}}_{ii}+1)+ t\boldsymbol{\mathcal{K}}_{ii} \nonumber
\end{aligned}
\end{equation}
Since $\det \mathbf{M}(t) \neq 0$ for all $t \in [0,1]$, $\mathbf{M}(t)$ is non-singular. Furthermore, $\det \mathbf{M}(0) = \prod_{i=1}^N (\boldsymbol{\mathcal{K}}_{ii} + 1) > 0$, ensuring $\det \boldsymbol{\mathcal{K}} = \det \mathbf{M}(1) > 0$. Additionally, each principal minor of $\boldsymbol{\mathcal{K}}$ is symmetric, strictly diagonally dominant, and has non-negative diagonal elements, resulting in positive determinants. By Sylvester’s criterion \cite{gilbert1991positive}, we conclude that $\boldsymbol{\mathcal{K}}$ is positive semi-definite, verifying \textit{Theorem~\ref{theo:kernel}}.
\end{proof}

\subsection{Kernel Approximation}\label{appro}
Kernel methods impose significant computational costs, with a complexity of $O(N^2)$ for $N$ observations when computing kernel matrices. To address this challenge, we employ the Nyström method \cite{Kumar2013Sampling}, integrated with self-representation learning, to approximate the kernel matrix in a low-rank form.

A straightforward way for reducing computational complexity is to determine \textit{k} cluster centroids by sampling \textit{Q} points from the dataset. According to the principles of self-representation learning, these cluster centroids can be represented as linear combinations of all data points, residing in the subspace spanned by the dataset $\mathbf{X}$. To avoid computing the entire kernel matrix, the centroids are restricted to a smaller subspace $\widehat{\mathbf{X}} \subset \mathbf{X}$, which must meet two properties: (1) $\widehat{\mathbf{X}}$ should be small enough to ensure efficiency, and (2) its coverage should be wide enough to reduce the approximation error.

Based on the kernel learning principles outlined in Section \ref{kernel}, particularly Eq~\eqref{eq:kernel} and \eqref{weight}, we can rapidly identify preliminary clusters or subspaces using the initial self representation matrix. Once these subspaces are identified, cluster centroids and surrounding data points can be selected from these subspaces to construct $\widehat{\mathbf{X}}$.

Given $\widehat{\mathbf{X}}$, which selects $Q$ data points $(Q \ll N)$, we define two kernel matrices: $\widehat{\boldsymbol{\mathcal{K}}} \in \mathcal{R}^{Q \times Q}$, representing the similarity among the selected points, and $\boldsymbol{\widetilde{\mathcal{K}}} \in \mathcal{R}^{N \times Q}$, capturing the similarity between the entire dataset and the selected points. Using the Nyström method, the full kernel matrix $\boldsymbol{\mathcal{K}}$ is approximated as $\boldsymbol{\mathcal{K}}^* \approx \boldsymbol{\widetilde{\mathcal{K}}} \widehat{\boldsymbol{\mathcal{K}}}^{-1} \boldsymbol{\widetilde{\mathcal{K}}}^\mathrm{T}$. Since $\widehat{\boldsymbol{\mathcal{K}}}$ is a subset of $\boldsymbol{\widetilde{\mathcal{K}}}$, only $\boldsymbol{\widetilde{\mathcal{K}}}$ needs to be computed, substantially reducing computational complexity compared to directly calculating $\boldsymbol{\mathcal{K}}$. To ensure $\boldsymbol{\mathcal{K}}$ remains positive semi-definite, a diagonal shift is applied to $\boldsymbol{\widetilde{\mathcal{K}}} \widehat{\boldsymbol{\mathcal{K}}}^{-1} \boldsymbol{\widetilde{\mathcal{K}}}^\mathrm{T}$ by adding a sufficiently large positive constant $\rho$, following the approach of Roth et al. \cite{roth2003optimal}. The steps for constructing the approximate kernel matrix $\boldsymbol{\mathcal{K}}^*$ are detailed in Algorithm~\ref{al1}.
\begin{algorithm}[ht]
\caption{Kernel Approximation}
\label{al1}
\LinesNumbered 
{\small{
\KwIn{$ \mathbf{X}, \mathbf{Z}, Q, \rho$}
\KwOut{$\boldsymbol{\mathcal{K}}^\ast\in \mathcal{R}^{N\times N}$}

Construct the initial self-representation matrix $\mathbf{Z}$ from $\mathbf{X}$\;
Identify preliminary subspaces based on $\mathbf{Z}$\;
Sample $Q$ points from these subspaces\;
Compute the kernel matrix $\boldsymbol{\widetilde{\mathcal{K}}}$ using Eq~\eqref{eq:kernel}, capturing the similarity between all points in $\mathbf{X}$ and the $Q$ sampled points\;
Extract the sub-matrix $\widehat{\boldsymbol{\mathcal{K}}}$, representing the similarity among the $Q$ selected points\;

Approximate the full kernel matrix $\boldsymbol{\mathcal{K}}$ through the Nyström method: $\boldsymbol{\mathcal{K}^\ast} \approx \boldsymbol{\widetilde{\mathcal{K}}}\widehat{\boldsymbol{\mathcal{K}}}^{{-1}}\boldsymbol{\widetilde{\mathcal{K}}}^{\mathrm{T}}+\rho\mathbf{I}$\;	
}}
\end{algorithm}
\begin{theorem}
The approximate kernel matrix $\rm\boldsymbol{\mathcal{K}^\ast}=(\boldsymbol{\widetilde{\mathcal{K}}}\widehat{\boldsymbol{\mathcal{K}}}^{{-1}}\boldsymbol{\widetilde{\mathcal{K}}}^{\mathrm{T}}+\rho\mathbf{I})$ is positive semi-definite.
\end{theorem}

\begin{proof}
The term $\boldsymbol{\widetilde{\mathcal{K}}} \widehat{\boldsymbol{\mathcal{K}}}^{-1} \boldsymbol{\widetilde{\mathcal{K}}}^\mathrm{T}$ is symmetric, and the diagonal shift ensures that $\boldsymbol{\mathcal{K}^\ast}$ becomes a symmetric, strictly diagonally dominant matrix, satisfying positive semi-definiteness.
\end{proof}

\subsection{Kernel Subspace Clustering Algorithm}
Our method follows the general procedure outlined in prior studies \cite{elhamifar2013sparse,lu2018subspace,bai2020sparse}. We first map the data $\mathbf{X}$ into a high-dimensional space using the learned kernel, then solve problem \eqref{eq:obj} to obtain the representation matrix $\mathbf{Z}$ in this nonlinear space. The affinity matrix is constructed as $(\mathbf{Z} + \mathbf{Z}^\mathrm{T}) / 2$. The final clustering results are obtained by applying spectral clustering to this affinity matrix.

\section{Theoretical Analysis of DKLM}
This section provides a comprehensive theoretical analysis of DKLM, detailing its underlying kernel and regularization schemes, the multiplicative triangle inequality of the learned kernel, and its connections to various existing clustering algorithms.
\subsection{Data-driven Kernel Representation Learning}
\subsubsection{Representation Matrix}
The core of DKLM is the kernel representation matrix $\mathbf{Z}$, which encapsulates the relationships and subspaces within the data. Without loss of generality, assume that $\mathbf{X} = [\mathbf{X}^{(1)}, \cdots, \mathbf{X}^{(k)}]$ is organized based on their respective subspace, the goal is to learn the representation $\mathbf{Z}$ where each data point is expressed as a combination of points from the same subspace. Specifically, for subspace $\mathbf{X}^{(i)}$, this can be written as $\mathbf{X}^{(i)} = \mathbf{X}^{(i)} \mathbf{Z}^{(i)}$. In such an idea case, the matrix $\mathbf{Z}$ in Eq.~\eqref{eq:subclustering} exhibits a $k$-block diagonal structure (up to permutations).
\begin{equation}
\begin{aligned}
\mathbf{Z}=&
\begin{bmatrix}

\mathbf{Z}^{(1)} &0& \cdots & 0 \\
0&\mathbf{Z}^{(2)} &\cdots&0\\

\vdots &\vdots & \ddots & \vdots \\

0 &0& \cdots & \mathbf{Z}^{(k)} 

\end{bmatrix}
\end{aligned}
\label{eq:sr}
\end{equation}
This representation reveals the underlying structure of $\mathbf{X}$, with each block $\mathbf{Z}^{(i)}$ in the diagonal representing a specific subspace. $k$ represents the number of blocks, which is directly associated with the number of distinct subspaces. For simplicity, we assume $\mathbf{X}$ is arranged based on true membership. However, in problem \eqref{eq:KernelSub}, the input matrix can also be expressed as $\tilde{\mathbf{X}} = \mathbf{XP}$, where $\mathbf{P}$ represents a permutation matrix that rearranges the columns of $\mathbf{X}$ (see \textit{Theorem~\ref{P-kernel}}).

It is also worth noting that $||\mathbf{Z}||_{\boxed{\scriptstyle k}}$ in \eqref{eq:kbdr} corresponds to $rank(\mathbf{L}_{\mathbf{Z}}) = N-k$. While one could consider utilizing $rank(\mathbf{L}_{\mathbf{Z}})$ as the $k$-block diagonal regularizer, this approach is not feasible in practice. This is because the number of data points $N$ typically far exceeds the number of clusters $k$, leading to a high rank for $\mathbf{L}_{\mathbf{Z}}$. Moreover, this method cannot effectively regulate the desired number of blocks, a critical factor in subspace clustering.

\begin{theorem}
    In DKLM, the representation matrix obtained for a permuted input data is equivalent to the permutation transformed original representation matrix. Specifically, let $\mathbf{Z}$ be feasible to $\Phi(\mathbf{X})=\Phi(\mathbf{X})\mathbf{Z}$, then $\tilde{\mathbf{Z}}=\mathbf{P^{T}ZP}$ is feasible to $\Phi(\tilde{\mathbf{X}})=\Phi(\tilde{\mathbf{X}})\tilde{\mathbf{Z}}$. (See Appendix~\ref{proof2} for proof.)
\label{P-kernel}
\end{theorem}
\subsubsection{Differences from Traditional Kernel Learning}
The main difference between a pre-defined kernel and our learned kernel lies in their underlying assumptions about the data. Pre-defined kernels typically presume that the data follows a specific distribution, such as Gaussian distribution, which may not hold true for heterogeneous real-world datasets. In contrast, our model learns the kernel directly from the data, creating a low-rank or sparse kernel matrix without such assumptions.

Moreover, our method is flexible and can be extended to multiple kernel clustering. Different self-representation regularizations yield various representation matrices, which can then be used to derive distinct kernels. For instance, low-rank self-representation captures the data’s low-rank structure, the $L_1$-norm induces sparsity, and the $L_{2,1}$-norm encourages row-sparse representations to reduce noise impact. These diverse kernels can be integrated into a consensus kernel that better captures the data’s underlying structure compared to traditional multiple kernel learning, which still relies on predefined kernel functions and assumptions about data distributions.

\subsection{Multiplicative Triangle Inequality of Kernel}
In all distance measures \( d(\cdot) \), the triangle inequality must be satisfied, i.e., \( d(\mathbf{X}_i, \mathbf{X}_l) + d(\mathbf{X}_l, \mathbf{X}_j) \geq d(\mathbf{X}_i, \mathbf{X}_j) \). Attempting to express this in terms of similarities as \( s(\mathbf{X}_i, \mathbf{X}_l) + s(\mathbf{X}_l, \mathbf{X}_j) \leq s(\mathbf{X}_i, \mathbf{X}_j) \) only holds if similarity is defined as $ s(\cdot) = -\ dissimilarity(\cdot) = -\ d(\cdot)$, which is problematic. While dissimilarities align naturally with distances with distances, similarities—especially in kernels—are typically positive as well. Our learned kernel adheres to a modified triangle inequality, known as the multiplicative triangle inequality, expressed as \( \boldsymbol{\mathcal{K}}(\mathbf{X}_i, \mathbf{X}_j) \geq \boldsymbol{\mathcal{K}}(\mathbf{X}_i, \mathbf{X}_l) \cdot \boldsymbol{\mathcal{K}}(\mathbf{X}_l, \mathbf{X}_j) \). This implies that the similarity between two points \( \mathbf{X}_i \) and \( \mathbf{X}_j \) is bounded by the product of their similarities to a third point \( \mathbf{X}_l \)  - \eg if \( \mathbf{X}_i \) and \( \mathbf{X}_j \) belong to the same subspace while \( \mathbf{X}_l \) is in a different subspace, the similarity between \( \mathbf{X}_i \) and \( \mathbf{X}_j \) is bounded below by the product of their similarities with \( \mathbf{X}_l \). This ensures that intra-subspace similarities are not underestimated, even when an intermediary point from a different subspace is considered, thereby preserving the subspace’s structure in the kernel (see the \textit{Theorem~\ref{theo:triangle}}).

\begin{theorem}\label{theo:triangle}
In our model, the learned kernel $\boldsymbol{\mathcal{K}}$ satisfies a multiplicative  triangle inequality, \ie
\begin{equation}
\boldsymbol{\mathcal{K}}(\mathbf{X}_i,\mathbf{X}_j) \geq \boldsymbol{\mathcal{K}}(\mathbf{X}_i,\mathbf{X}_l)\cdot\boldsymbol{\mathcal{K}}(\mathbf{X}_l,\mathbf{X}_j)
\label{eq:mti}
\nonumber
\end{equation}
\end{theorem}
\begin{proof}
Let $\mathbf{H}_{ij}= 2\cdot max(\mathbf{G})-\mathbf{G}_{ij}$. Due to the fact that $2\cdot max(\mathbf{G})-\mathbf{G}_{il}-\mathbf{G}_{lj}+\mathbf{G}_{ij}\geq 0$, we have
\begin{equation}
\begin{gathered}
    \mathbf{H}_{ij} \leq \mathbf{H}_{il} + \mathbf{H}_{lj}\\
    \Updownarrow \\
        \mathrm{exp}(-\mathbf{H}_{ij})\geq \mathrm{exp}(-\mathbf{H}_{il}-\mathbf{H}_{lj})\\
    \Updownarrow\\
    \mathrm{exp}(-\mathbf{H}_{ij})\geq \mathrm{exp}(-\mathbf{H}_{il})\cdot \mathrm{exp}(-\mathbf{H}_{lj}) \Leftrightarrow \boldsymbol{\mathcal{K}}_{ij} \geq \boldsymbol{\mathcal{K}}_{il}\cdot \boldsymbol{\mathcal{K}}_{lj}
\end{gathered}
\nonumber
\end{equation}
\end{proof}

\subsection{Connection to Clustering Algorithms}
In this subsection, we will go over the conditions under which various clustering techniques are equivalent to the objective function \eqref{eq:obj}. 
\begin{theorem}\label{theo:kkmeans}
When $\alpha =2$ and $\gamma \rightarrow \infty$, \eqref{eq:obj} is equivalent to kernel $k$-means \cite{dhillon2004kernel}.
\end{theorem}
\begin{proof}
When $\gamma \rightarrow \infty$, $\min_{\mathbf{Z}} \gamma ||\mathbf{Z}||_{\boxed{\scriptstyle k}}$ will ensures $||\mathbf{Z}||_{\boxed{\scriptstyle k}} \rightarrow 0$. Thus, combining this with $\alpha =2$, \eqref{eq:obj} becomes
\begin{equation}
\begin{aligned}
         \min_{\mathbf{Z}} \frac{1}{2}\mathrm{Tr}(\boldsymbol{\mathcal{K}}+&\mathbf{Z}^\mathrm{T}\boldsymbol{\mathcal{K}}\mathbf{Z})-2\mathrm{Tr}(\boldsymbol{\mathcal{K}}\mathbf{Z})=\frac{1}{2}\mathrm{Tr}(\boldsymbol{\mathcal{K}}-2\boldsymbol{\mathcal{K}}\mathbf{Z}+\mathbf{Z}^\mathrm{T}\boldsymbol{\mathcal{K}}\mathbf{Z}) \\ &s.t.\ \mathbf{Z} = \mathbf{Z}^\mathrm{T} \geq 0, \mathrm{diag}(\mathbf{Z})=0, \mathbf{{1}^{\mathrm{T}}\mathbf{Z}=\mathbf{1}^{\mathrm{T}}}
\end{aligned}
\label{eq:kkmeansmatrix}
\end{equation}
Let the solution set of this form be denoted as $\mathcal{C}$, hence the above problem can be reformulated as
\begin{equation}
    \min_{\mathbf{Z}_i \in \mathcal{C}} \sum_i ||\mathrm{\Phi}(\mathbf{X}_i)-\mathrm{\Phi}(\mathbf{X})\mathbf{Z}_i||^2
\label{eq:kkmeans}
\end{equation}
It can be easily inferred that the mean of the $i$-th cluster in the kernel space is $\mathrm{\Phi}(\mathbf{X})\mathbf{Z}_i$. Therefore, \eqref{eq:kkmeans} represents the kernel $k$-means.
\end{proof}
\begin{corollary}
When $\alpha =2$ and $\gamma \rightarrow \infty$, and a linear kernel is adopted, the \eqref{eq:obj} is equivalent to k-means.
\end{corollary}
\begin{proof}
It is evident when no transformation is applied to $\mathbf{X}$ in \eqref{eq:kkmeans}.
\end{proof}

\begin{theorem}
When $\alpha$ is sufficiently large, $Reg(\mathbf{Z}) = ||\mathbf{Z}||_*$, and a linear kernel is adopted, the \eqref{eq:obj} is equivalent to subspace clustering via latent low-rank representation \cite{liu2012robust}.
\end{theorem}
\begin{proof} When $\alpha$ is sufficiently large, the term $-\alpha \mathrm{Tr}(KZ)$ will dominate the objective function \eqref{eq:obj}, and thus $\mathbf{Z}$ will tend to be close to $\boldsymbol{\mathcal{K}}$. Since $\boldsymbol{\mathcal{K}}$ is symmetric and positive semi-definite, it can be decomposed as $\boldsymbol{\mathcal{K}} = \mathbf{U}\Tilde{\Sigma} \mathbf{U}^{\mathrm{T}}$, where $\mathbf{U}$ is an orthogonal matrix and $\Tilde{\Sigma}$ is a diagonal matrix with nonnegative entries. Therefore, $\mathbf{Z}$ can be approximated by $\mathbf{Z} \approx \mathbf{U}\Tilde{\Sigma} \mathbf{U}^{\mathrm{T}}$, where $\Tilde{\Sigma}$ is a diagonal matrix with some entries of $\sum$ set to zero. This means that $\mathbf{Z}$ has a low-rank structure, which can be encouraged by using the nuclear norm $||\mathbf{Z}||_*$ as the regularizer. When a linear kernel is adopted, $\boldsymbol{\mathcal{K}} = \mathbf{X}^{\mathrm{T}}\mathbf{X}$, and thus $\mathbf{Z} \approx \mathbf{X}^{\mathrm{T}}\Tilde{\mathbf{X}}$ for some low-rank matrix $\Tilde{\mathbf{X}}$. This is equivalent to the low-rank representation model proposed in \cite{liu2012robust}.
\end{proof}

\begin{theorem}
When $\alpha=2$ and $Reg(\mathbf{Z}) = ||\mathbf{Z}||_E$, the \eqref{eq:obj} is equivalent to subspace clustering with entropy norm (SSCE) \cite{bai2020sparse}.
\end{theorem}
\begin{proof}
When $\alpha = 2$, the \eqref{eq:obj} becomes
\begin{equation}
\min_{\mathbf{Z}} \frac{1}{2}\mathrm{Tr}(\boldsymbol{\mathcal{K}} + \mathbf{Z}^\mathrm{T}\boldsymbol{\mathcal{K}}\mathbf{Z}) - 2\mathrm{Tr}(\boldsymbol{\mathcal{K}}\mathbf{Z}) + \gamma||\mathbf{Z}||_E
\end{equation}
where $||\mathbf{Z}||_E = -\mathrm{Tr}(\mathbf{Z}\log \mathbf{Z})$. This is exactly the subspace clustering with entropy model proposed by Bai et al. \cite{bai2020sparse}, which uses the entropy norm to encourage diversity on $\mathbf{Z}$.
\end{proof}

\begin{theorem}\label{theo:KNN}
When $\alpha$ is sufficiently large, $\mathrm{Reg}(\mathbf{Z})= ||\mathbf{Z}||^2$, and a linear kernel is adopted, the \eqref{eq:obj} is equivalent to subspace clustering via good neighbors \cite{yang2019subspace}.
\end{theorem}
\begin{proof}
When $\alpha$ is sufficiently large, if $\mathbf{X}_i$ and $\mathbf{X}_j$ are far apart (\ie, $\boldsymbol{\mathcal{K}}(\mathbf{X}_i, \mathbf{X}_j)$ is small), it can be seen that from \eqref{eq:kz} the $\mathbf{Z}_{ij}$ tends to 0. In other words, for $\mathbf{X}_i$, the self-representation coefficient is not 0 only ffor points similar to it or within its nearest neighbors. So the \eqref{eq:obj} can be transformed to
\begin{equation}
    \min_{\mathbf{Z}}\frac{1}{2} \sum_i || \mathrm{\Phi}(\mathbf{X}_i)-\sum_{\mathbf{X}_j \in N_k(\mathbf{X}_i)}\mathrm{\Phi}(\mathbf{X}_j)\mathbf{Z}||^2+\gamma||\mathbf{Z}||^2
\end{equation}
where $N_k(\mathbf{X}_i)$ is a set of $k$ nearest neighbors of $\mathbf{X}_i$. It can be seen that it is equivalent to subspace clustering via good neighbors when a linear kernel is adopted.
\end{proof}

\section{Experiments}
This section presents a series of experiments designed to evaluate the performance and validate the effectiveness of DKLM. We compare its performance against five self-representation-based methods: SSC \cite{elhamifar2013sparse}, LRR \cite{liu2012robust}, BDR \cite{lu2018subspace}, SSCE \cite{bai2020sparse}, and LSR \cite{lu2012robust}. Additionally, we include four kernel-based approaches in the comparison: spectral clustering (SC) \cite{ng2001spectral}, kernel SSC (KSSC) \cite{patel2014kernel}, kernel $k$-means (KKM) \cite{scholkopf1998nonlinear}, and robust kernel $k$-means (RKKM) \cite{du2015robust}. The implementations of the baseline methods are based on the publicly available code provided by their respective authors. To evaluate clustering performance, we use three commonly employed external evaluation metrics: clustering accuracy (ACC), normalized mutual information (NMI), and purity. These metrics quantify the similarity between the clustering results and the ground truth, where higher values indicate better clustering quality. In this work, LRR is primarily adopted to facilitate learning a kernel with low-rank feature mapping in Hilbert space. For all experiments, the sampling size $Q$ is set to one-third of the total data points.

\subsection{Experiments on synthetic data}
\begin{figure}[t]
\centerline{\includegraphics[width=0.9\linewidth]{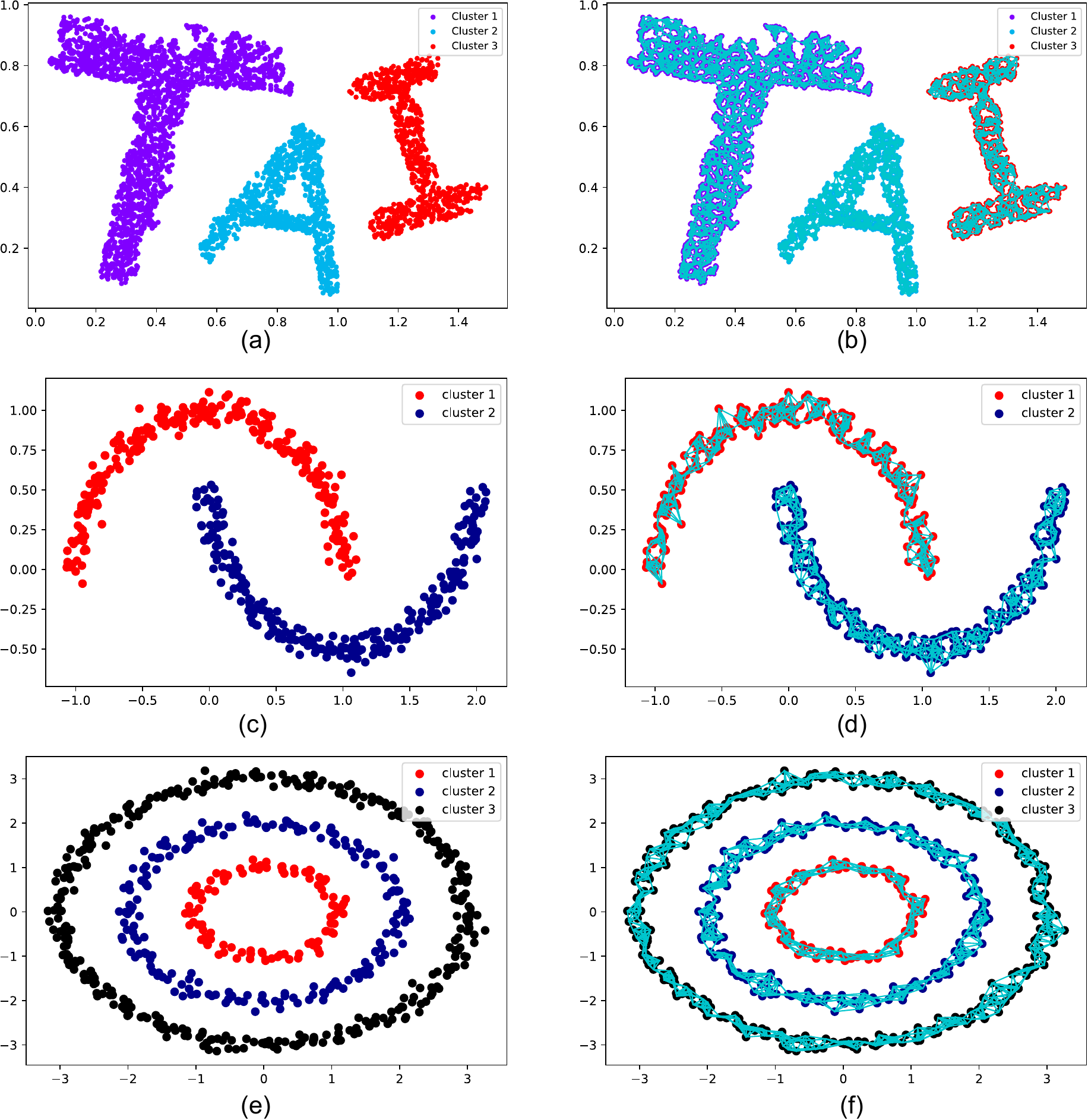}}
\caption{Visualization of results on three 2D datasets: (a), (c), and (e) represent the three synthetic datasets, while (b), (d), and (f) show the affinity graphs $\mathbf{Z}$ learned by the model. For clarity, only edges with weights of 0.001 or higher are displayed. }
\label{Fig:toy1}
\end{figure}

We generated three synthetic datasets to visually illustrate the capability of DKLM and conducted experiments on them as follows:

1) \textit{2D Scenarios}: The first dataset (SyD1) comprises two clusters of 2000 samples arranged in the shapes of the letters 'T', 'A' and 'I', as shown in Fig.\ref{Fig:toy1}(a). The second dataset (SyD2), referred to as the \textit{Two-Moon dataset}, contains two clusters of 500 points each, forming the upper (red) and lower (blue) moons, as depicted in Fig.\ref{Fig:toy1}(c). The third dataset (SyD3), called the \textit{Three-Ring dataset}, has three clusters with 650 points each, as illustrated in Fig.\ref{Fig:toy1}(e). For these datasets, using LRR to learn a kernel is ineffective because the ambient and intrinsic data dimensions are equal. Instead, SSC is employed. As shown in Fig.~\ref{Fig:toy1}(b), (d), and (f), DKLM effectively separates clusters by learning an affinity graph $\mathbf{Z}$ where intra-cluster connections are strong, and inter-cluster connections are absent. By preserving local manifold structures, DKLM identifies the true cluster structure and the correct number of connected components in the graph.

2) \textit{High-Dimensional Scenarios:}  A synthetic dataset (SyD4) in 78 dimensions with 1300 samples distributed across four clusters was created. These clusters were generated using four nonlinear functions, with 20\% of the data vectors corrupted by random noise. 
\begin{equation}
    \begin{cases} 
    \begin{aligned}
       f_1(d) &= \cos\left(4\pi d / 7\right) + \cos\left(\pi(d - 40)\right) \\
       f_2(d) &= \sin\left(\pi d / 4 - 4\right) - \sin\left(\pi d / 5\right) \\
       f_3(d) &= 1 - \sin\left(\pi d / 3\right) \times \cos\left(\pi(d - 4) / 5\right) \times \cos\left(\pi d\right) \\
       f_4(d) &= \sin\left(\pi d / 3\right) \times \cos\left(\pi d / 6\right) \times \cos\left(\pi(d - 12)\right)
    \end{aligned}
    \end{cases}
   \label{eq:sydeq}
   \nonumber
\end{equation}

\begin{figure}[t]
\centerline{\includegraphics[width=1\linewidth]{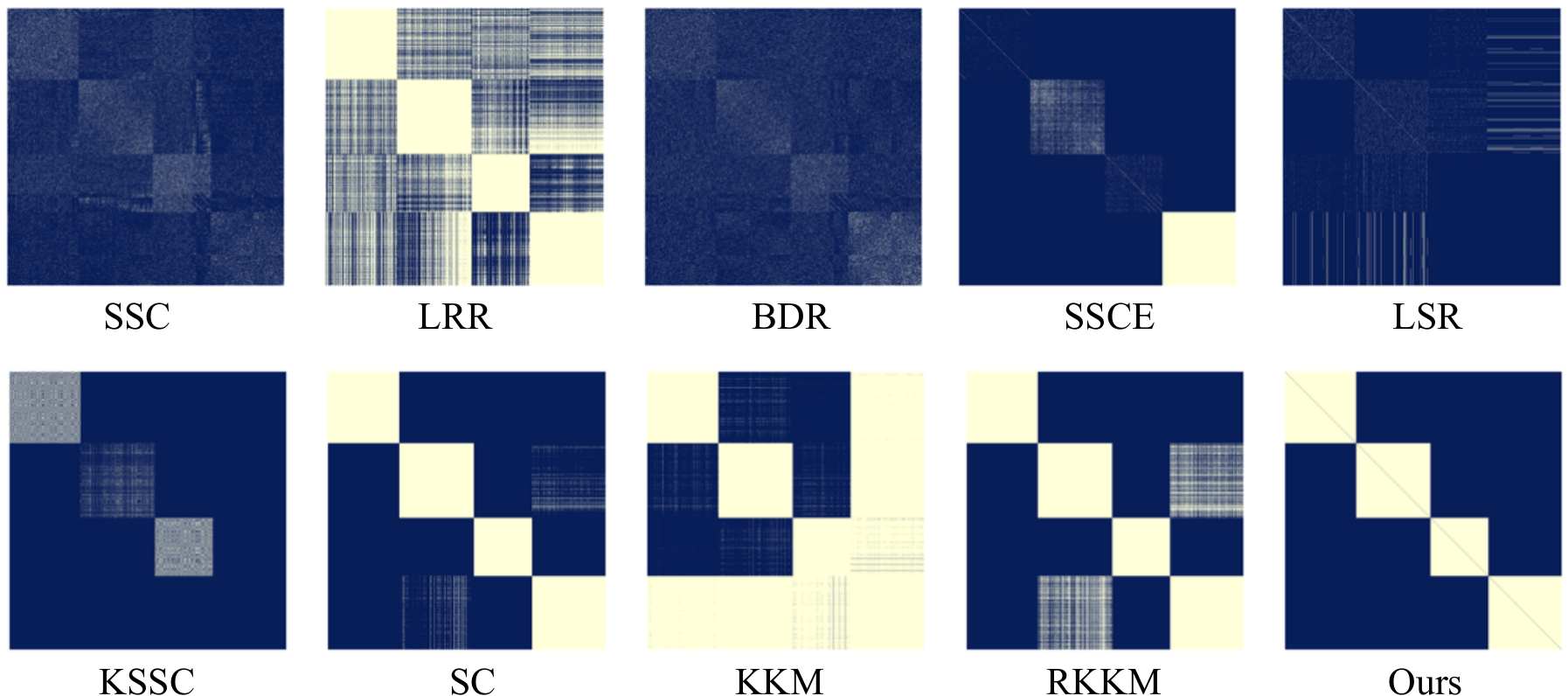}}
\caption{Binarized affinity matrices on high-dimensional SyD4, comparing various SOTA methods with ours.}
\label{Fig:toy2}
\end{figure}    

\begin{figure*}[t]
\centerline{\includegraphics[width=1\linewidth]{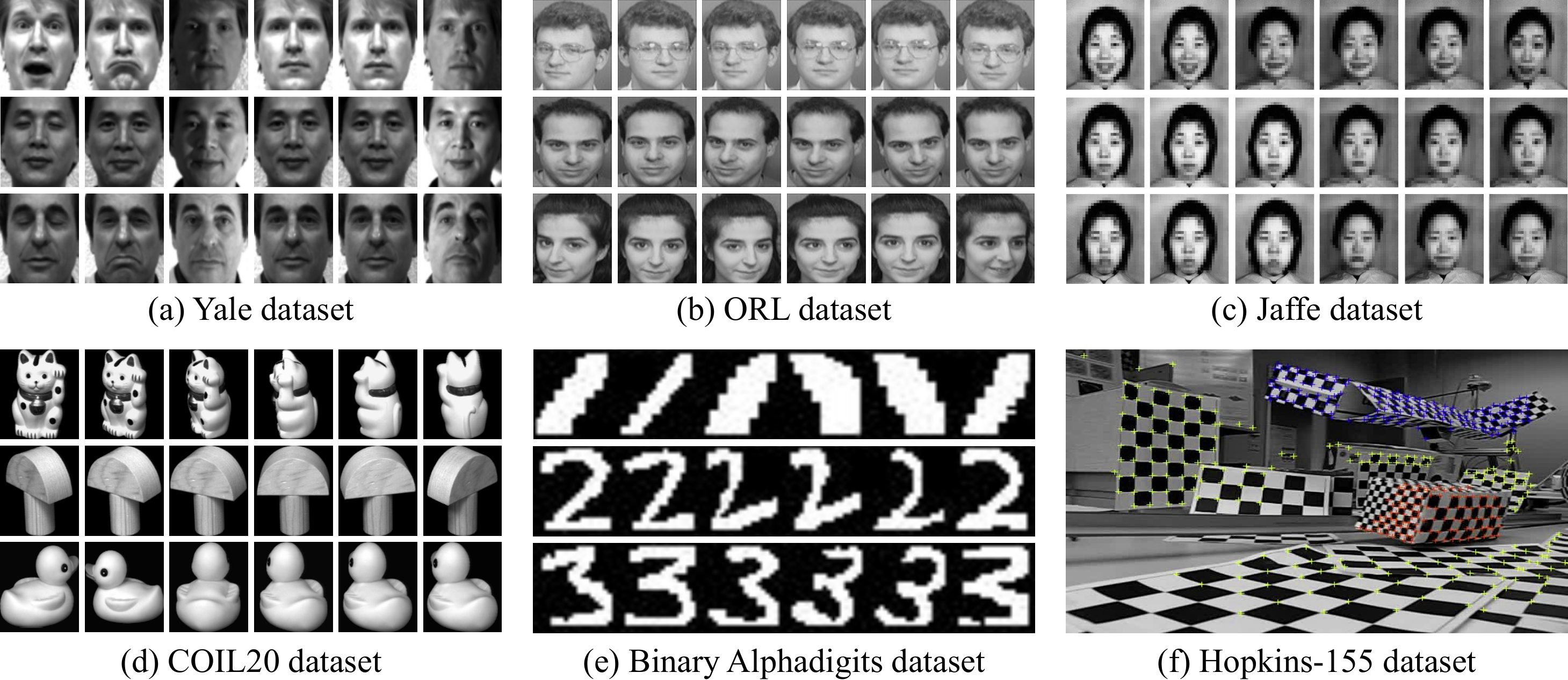}}
\caption{Sample images from Yale, ORL, Jaffe, COIL20, Binary Alphadigits, and Hopkins-155 datasets}
\label{Fig:realdata}
\end{figure*}

Figure \ref{Fig:toy2} shows a comparison of the binarized representation matrices generated by DKLM and other state-of-the-art methods. DKLM effectively generates a block-diagonal matrix characterized by dense intra-cluster connections and sparse inter-cluster separations, distinctly illustrating the cluster structure. In contrast, SSC, BDR, SSCE, and LSR struggle to differentiate clusters in scenarios involving overlapping or nonlinear subspaces. While LRR shows improved performance for weakly correlated points, it fails to distinguish strongly correlated points, leading to indistinct clusters. Kernel-based methods, though capable of handling nonlinearities, face challenges when dealing with local manifold structures. For instance, KSSC identifies only three clusters, but SC, KKM, and RKKM are unable to differentiate between the second and fourth clusters. These shortcomings may stem from the reliance on pre-defined kernel functions. Instead, the number of blocks in our approach corresponds to the number of nonlinear functions used to generate the data.
\subsection{Experiments on real-world data}
We evaluate our method using nine publicly available datasets, including five image datasets: Yale\footnote{\url{http://www.cvc.yale.edu/projects/yalefaces/yalefaces.html}}, ORL\footnote{\url{https://www.cl.cam.ac.uk/research/dtg/attarchive/facedataset.html}}, Jaffe\footnote{\url{http://www.kasrl.org/jaffe.html}}, COIL20\footnote{\url{http://www.cs.columbia.edu/CAVE/software/softlib/coil-20.php}}, and Binary Alphadigits\footnote{\url{https://cs.nyu.edu/roweis/data/binaryalphadigs.mat.}}. Additionally, three text corpora – TR11, TR41, and TR45 – and the Hopkins 155 database\footnote{\url{http://www.vision.jhu.edu/data/hopkins155/}} for motion segmentation are included. Detailed statistics and dimensionalities of these datasets are summarized in Table~\ref{tab:data}.
\begin{table}[h]

\centering
\caption{Summaries of the data sets}
\label{tab:data}
\begin{threeparttable} 
\resizebox{0.5\textwidth}{!}{
\begin{tabular}{cccc}
\hline
   Datasets         & \#instances & \#features & \#clusters \\ \hline
Yale        & 165         & 1024 (32 $\times$ 32)      & 15         \\
ORL         & 400         & 1024 (32 $\times$ 32)       & 40         \\
Jaffe       & 213         & 676  (26 $\times$ 26)       & 10         \\
COIL20      & 1440        & 1024 (32 $\times$ 32)      & 20         \\
BA          & 1404        & 320 (20 $\times$ 16)       & 36         \\
Hopkins 155\tnote{1} & --           & --          & --         \\
TR11        & 414         & 6429       & 9          \\
TR41        & 878         & 7454       & 10         \\
TR45        & 690         & 8261       & 10         \\\hline
\end{tabular}}

\begin{tablenotes}    
        \footnotesize     
        \item[1] Hopkins 155 is a database consisting of 155 sequences data set. For more details, refer to Section~\ref{sec:hop}.
\end{tablenotes} 
\end{threeparttable} 
\end{table}

\begin{table*}[!t]
    \centering
    \caption{Comparison results with different methods. The best results are highlighted in \textcolor{red}{\textbf{bold}} and the second best are \textcolor{blue}{\underline{underlined}}.}
    \label{Tab:extended}
    
    \subfloat[The comparison result of self-representation-based methods.\label{Tab:extended1}]{
        \resizebox{0.85\textwidth}{!}
{
        \begin{tabular}{ccccccc}
            \hline
Dataset                             & Metrics & SSC            & LRR            & BDR            & SSCE                                     & LSR            \\ \hline
\multirow{3}{*}{Yale}               & ACC     & 0.3988(0.0434) & 0.4857(0.0370) & \textcolor{blue}{\underline{0.4951(0.0214)}} & 0.4668(0.0273)                           & 0.4710(0.0325) \\
                                    & NMI     & 0.4375(0.0348) & \textcolor{blue}{\underline{0.5432(0.0316)}} & 0.5332(0.0205) & 0.5235(0.0311)                           & 0.5246(0.0239) \\
                                    & Purity  & 0.4149(0.0382) & 0.5019(0.0348) & 0.5149(0.0248) & 0.4829(0.0297)                           & 0.5030(0.0251) \\ \hline
\multirow{3}{*}{ORL}                & ACC     & 0.5717(0.0238) & 0.5312(0.0412) & 0.5804(0.0141) & \textcolor{blue}{\underline{0.6751(0.0251)}}                           & 0.5835(0.0416) \\
                                    & NMI     & 0.6872(0.0314) & 0.7338(0.0527) & 0.7524(0.0170) & \textcolor{red}{\pmb{0.8024(0.0241)}} & 0.7623(0.0315) \\
                                    & Purity  & 0.5939(0.0342) & 0.6586(0.0414) & 0.6857(0.0129) & \textcolor{blue}{\underline{0.7443(0.0220)}}                           & 0.6385(0.0430) \\ \hline
\multirow{3}{*}{Jaffe}              & ACC     & 0.7431(0.0530) & 0.7481(0.0149) & 0.7946(0.0527) & 0.7742(0.0624)                           & 0.7216(0.0595) \\
                                    & NMI     & 0.7984(0.0483) & 0.8147(0.0250) & 0.8486(0.0602) & 0.8299(0.0570)                           & \textcolor{blue}{\underline{0.8794(0.0531)}} \\
                                    & Purity  & 0.7739(0.0415) & 0.7883(0.0273) & \textcolor{blue}{\underline{0.8195(0.0530)}} & 0.7846(0.0519)                           & 0.7631(0.0461) \\ \hline
\multirow{3}{*}{COIL20}             & ACC     & 0.7319(0.0120) & 0.7737(0.0563) & \textcolor{blue}{\underline{0.8066(0.0499)}} & 0.7836(0.0269)                           & 0.6344(0.0322) \\
                                    & NMI     & \textcolor{blue}{\underline{0.8903(0.0801)}} & 0.8816(0.0327) & \textcolor{red}{\pmb{0.8959(0.0246)}} & 0.8855(0.0144)                           & 0.7583(0.0390) \\
                                    & Purity  & 0.8042(0.1207) & 0.8573(0.0129) & 0.8543(0.0374) & \textcolor{blue}{\underline{0.8626(0.0138)}}                           & 0.6187(0.0106) \\ \hline
\multirow{3}{*}{BA} & ACC     & 0.2422(0.0130) & 0.4537(0.0427) & \textcolor{blue}{\underline{0.5084(0.0135)}} & 0.4628(0.0179)                           & 0.2473(0.0282) \\
                                    & NMI     & 0.3741(0.0203) & 0.5797(0.0431) & 0.5547(0.0149) & \textcolor{blue}{\underline{0.5607(0.0053)}}                           & 0.3653(0.0029) \\
                                    & Purity  & 0.3041(0.0104) & 0.5329(0.0424) & \textcolor{blue}{\underline{0.5738(0.0257)}} & 0.5424(0.0034)                           & 0.2854(0.0036) \\ \hline
\multirow{3}{*}{TR11}               & ACC     & 0.3261(0.0237) & 0.3867(0.0418) & 0.3013(0.0150) & 0.3837(0.0138)                           & 0.4096(0.0111) \\
                                    & NMI     & 0.3641(0.0105) & 0.3061(0.0212) & 0.3506(0.0242) & 0.3207(0.0104)                           & 0.3829(0.0132) \\
                                    & Purity  & 0.3314(0.0112) & 0.3519(0.0120) & 0.3634(0.0207) & 0.3603(0.0153)                           & 0.3535(0.0228) \\ \hline
\multirow{3}{*}{TR41}               & ACC     & 0.2804(0.0187) & 0.4062(0.0138) & 0.4274(0.0291) & 0.3958(0.0093)                           & 0.3461(0.0228) \\
                                    & NMI     & 0.2308(0.0035) & 0.3750(0.0322) & \textcolor{blue}{\underline{0.3857(0.0135)}} & 0.4009(0.0315)                           & 0.3004(0.0196) \\
                                    & Purity  & 0.2531(0.0214) & 0.3542(0.0484) & 0.3913(0.0065) & 0.3712(0.0043)                           & 0.2935(0.0415) \\ \hline
\multirow{3}{*}{TR45}               & ACC     & 0.2438(0.0014) & 0.3984(0.0239) & 0.4123(0.0184) & 0.3890(0.0217)                           & 0.3284(0.0159) \\
                                    & NMI     & 0.1684(0.0013) & \textcolor{blue}{\underline{0.4301(0.0024)}} & \textcolor{red}{\pmb{0.4421(0.0123)}}& 0.4156(0.0148)                           & 0.3789(0.0112) \\
                                    & Purity  & 0.2134(0.0229) & 0.4241(0.0034) & 0.4354(0.0167) & 0.4023(0.0134)                           & 0.2654(0.0193)  \\ \hline
        \end{tabular}}
    }
    
    \vspace{0em} 
 
    \subfloat[The comparison result of kernel-based methods.\label{Tab:extended2}]{
        \resizebox{0.85\textwidth}{!}
{
        \begin{tabular}{ccccccc}
            \hline
Dataset                             & Metrics & KSSC           & SC             & KKM            & RKKM           & DKLM                                     \\ \hline
\multirow{3}{*}{Yale}               & ACC     & 0.4538(0.0340) & 0.4920(0.0241) & 0.4742(0.0427) & 0.4853(0.0415) & \textcolor{red}{\pmb{0.5850(0.0153)}} \\
                                    & NMI     & 0.4993(0.0298) & 0.5218(0.0210) & 0.5174(0.0482) & 0.5241(0.0371) & \textcolor{red}{\pmb{0.6441(0.0142)}} \\
                                    & Purity  & \textcolor{blue}{\underline{0.5319(0.0281)}} & 0.5207(0.0109) & 0.4855(0.0419) & 0.4929(0.0314) & \textcolor{red}{\pmb{0.6204(0.0081)}}\\ \hline
\multirow{3}{*}{ORL}                & ACC     & 0.5880(0.0310) & 0.5792(0.0351) & 0.5319(0.0287) & 0.5576(0.0320) & \textcolor{red}{\pmb{0.6862(0.0215)}} \\
                                    & NMI     & 0.7689(0.0251) & 0.7618(0.0193) & 0.7415(0.0312) & 0.7343(0.0286) & \textcolor{blue}{\underline{0.7865(0.0191)}}                           \\
                                    & Purity  & 0.6846(0.0153) & 0.6350(0.0329) & 0.5721(0.0284) & 0.5960(0.0347) & \textcolor{red}{\pmb{0.7502(0.0249)}} \\ \hline
\multirow{3}{*}{Jaffe}              & ACC     & \textcolor{blue}{\underline{0.8181(0.0841)}} & 0.7863(0.0457) & 0.7549(0.0810) & 0.7651(0.0763) & \textcolor{red}{\pmb{0.9831(0.0023)}}\\
                                    & NMI     & 0.7923(0.0665) & 0.8341(0.0381) & 0.8037(0.0731) & 0.8437(0.0715) & \textcolor{red}{\pmb{0.9618(0.0017)}} \\
                                    & Purity  & 0.7484(0.0627) & 0.7438(0.0416) & 0.7824(0.0756) & 0.8059(0.0702) & \textcolor{red}{\pmb{0.9853(0.0021)}} \\ \hline
\multirow{3}{*}{COIL20}             & ACC     & 0.7535(0.0231) & 0.4390(0.0171) & 0.7439(0.0402) & 0.7792(0.0031) & \textcolor{red}{\pmb{0.8234(0.0242)}}                           \\
                                    & NMI     & 0.8843(0.0339) & 0.5381(0.0134) & 0.8372(0.0310) & 0.8382(0.0026) & 0.8344(0.0340)                           \\
                                    & Purity  & 0.7637(0.0310) & 0.4723(0.0249) & 0.7512(0.0238) & 0.7634(0.0032) & \textcolor{red}{\pmb{0.8938(0.0311)}}                           \\ \hline
\multirow{3}{*}{BA} & ACC     & 0.3513(0.0147) & 0.3641(0.0128) & 0.4341(0.0192) & 0.4446(0.0183) & \textcolor{red}{\pmb{0.5584(0.0132)}}                           \\
                                    & NMI     & 0.3832(0.0153) & 0.3912(0.0112) & 0.4349(0.0080) & 0.5012(0.0093) & \textcolor{red}{\pmb{0.6729(0.0124)}}                           \\
                                    & Purity  & 0.4123(0.0105) & 0.3923(0.0096) & 0.4214(0.0113) & 0.5054(0.0102) & \textcolor{red}{\pmb{0.5984(0.0097)}}                           \\ \hline
\multirow{3}{*}{TR11}               & ACC     & 0.4162(0.0312) & 0.4332(0.0293) & 0.4401(0.0683) & \textcolor{blue}{\underline{0.4502(0.0031)}} & \textcolor{red}{\pmb{0.4633(0.0323)}}                           \\
                                    & NMI     & 0.3246(0.0205) & 0.3145(0.0194) & 0.3324(0.0213) & 0.3425(0.0224) & \textcolor{red}{\pmb{0.4356(0.0235)}}                          \\
                                    & Purity  & 0.5239(0.0195) & 0.5138(0.0184) & \textcolor{red}{\pmb{0.5660(0.0163)}} & \textcolor{blue}{\underline{0.5354(0.0206)}} & \textcolor{red}{\pmb{0.5660(0.0217)}}                           \\ \hline
\multirow{3}{*}{TR41}               & ACC     & 0.4469(0.0334) & 0.4348(0.0021) & 0.4661(0.0363) & \textcolor{blue}{\underline{0.4694(0.0591)}} & \textcolor{red}{\pmb{0.4830(0.0345)}}                          \\
                                    & NMI     & 0.3864(0.0225) & 0.3763(0.0214) & 0.3942(0.0233) & \textcolor{blue}{\underline{0.4043(0.0244)}} & \textcolor{red}{\pmb{0.4074(0.0255)}}                          \\
                                    & Purity  & 0.5751(0.0325) & 0.5650(0.0314) & \textcolor{blue}{\underline{0.5852(0.0336)}} & \textcolor{red}{\pmb{0.5953(0.0347)}} & \textcolor{red}{\pmb{0.5964(0.0358)}}                           \\ \hline
\multirow{3}{*}{TR45}               & ACC     & \textcolor{blue}{\underline{0.4693(0.0153)}} & 0.4592(0.0142) & 0.4456(0.0345) & 0.4546(0.0394) &\textcolor{red}{\pmb{0.4727(0.0164)}}                           \\
                                    & NMI     & 0.3424(0.0139) & 0.3323(0.0128) & 0.3492(0.0147) & 0.3593(0.0158) & 0.3824(0.0169)                           \\
                                    & Purity  & 0.5122(0.0154) & 0.5021(0.0143) & 0.5190(0.0162) & \textcolor{blue}{\underline{0.5291(0.0173)}} & \textcolor{red}{\pmb{0.5422(0.0184)}}  \\ \hline
        \end{tabular}}
    }
\end{table*}

\begin{figure*}[t]
\centerline{\includegraphics[width=1\linewidth]{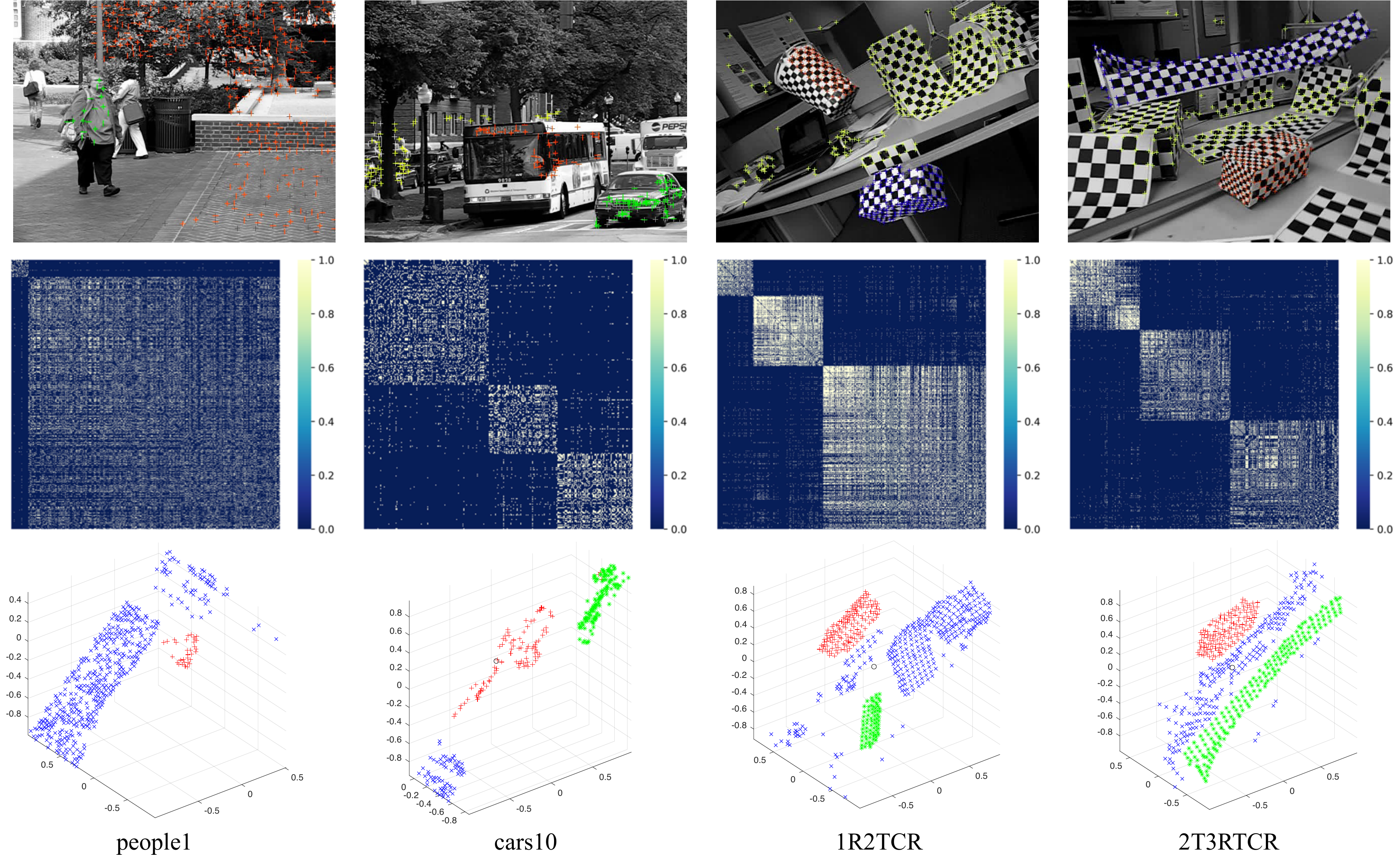}}
\caption{Clustering results for four sequences (people1, cars10, 1R2TCR, 2T3RTCR) from the Hopkins155 database. The top row shows sample images with overlaid tracked points, the second row displays self-representation matrices generated by DKLM, and the bottom row provides 3D visualizations of the clustering outcomes.}
\label{Fig:hop}
\end{figure*}
\subsubsection{Image and Text Clustering}

We test DKLM on image datasets and text datasets. These datasets present various challenges, from the complex nonlinear structures often found in image data to the high-dimensional, sparse nature of text data.

The Yale dataset comprises 165 grayscale images of faces from 15 individuals, with 11 images per subject taken under varying lighting conditions, facial expressions, and with or without glasses. The ORL dataset includes 400 face images from 40 individuals, each with 10 images captured at different times, showing variations in lighting and expressions. The Jaffe dataset contains 213 images of 10 Japanese female models, each portraying 7 distinct facial expressions. COIL20 consists of 1,440 grayscale images of 20 objects, with 72 images per object taken from multiple angles. The Binary Alphadigits dataset features 1,404 images representing digits (0-9) and letters (A-Z), each resized to 20x16 pixels. For text datasets, TR11, TR41, and TR45 are widely used for document clustering tasks, comprising 414, 878, and 690 documents, respectively, with features derived from term-frequency representations.

We conduct clustering experiments on these datasets using both self-representation and kernel methods. The results, as presented in Tables~\ref{Tab:extended}(a) and~\ref{Tab:extended}(b), demonstrate the superior performance of our proposed method, DKLM, especially on face datasets like Yale, ORL, and Jaffe, which exhibit strong nonlinear characteristics. These datasets are known to have complex structures due to variations in pose and illumination. Notably, under the Lambertian assumption \cite{basri2003lambertian}, face images of a subject taken under fixed poses and different lighting conditions approximately form a 9-dimensional linear subspace. Consequently, when combining images from $k$ subjects, the data is organized as a union of multiple 9-dimensional subspaces, adding to the nonlinear complexity.

It is worth noting that kernel-based models perform significantly better than self-representation-based methods on the ORL dataset, primarily due to the dataset’s pronounced nonlinear manifold structure, which kernel methods are well-suited to handle. For example, DKLM achieves an accuracy of 0.6862 on the ORL dataset, outperforming all other methods. Notably, SSCE also yields competitive results on this dataset, as SSC with entropy norm has been shown to be equivalent to direct spectral clustering under specific constraints \cite{bai2020sparse}. Similarly, on the Jaffe dataset, DKLM achieved an outstanding accuracy of 0.9831, clearly surpassing other approaches. These results highlight DKLM’s effectiveness in capturing the intricate structures within face datasets, making it especially suitable for tasks involving complex, nonlinear data. DKLM’s ability to learn kernels directly from the data, tailored to these underlying structures, allowed it to outperform traditional methods that rely on predefined kernels. Moreover, the fact that our kernel satisfies the multiplicative triangular inequality constraint, which guarantees a lower bound on similarity, further enhances DKLM’s performance, ensuring that the learned kernels better capture the true relationships within the data.

For text datasets, DKLM also showed good performance, particularly in its ability to capture the complex structure inherent in high-dimensional, sparse data like text. On the TR11, TR41, and TR45 datasets, DKLM consistently outperformed several other methods, achieving the best overall clustering results across these datasets. These results confirm DKLM’s robustness in handling the challenges of textual data.  

\begin{figure}[t]
\centerline{\includegraphics[width=1\linewidth]{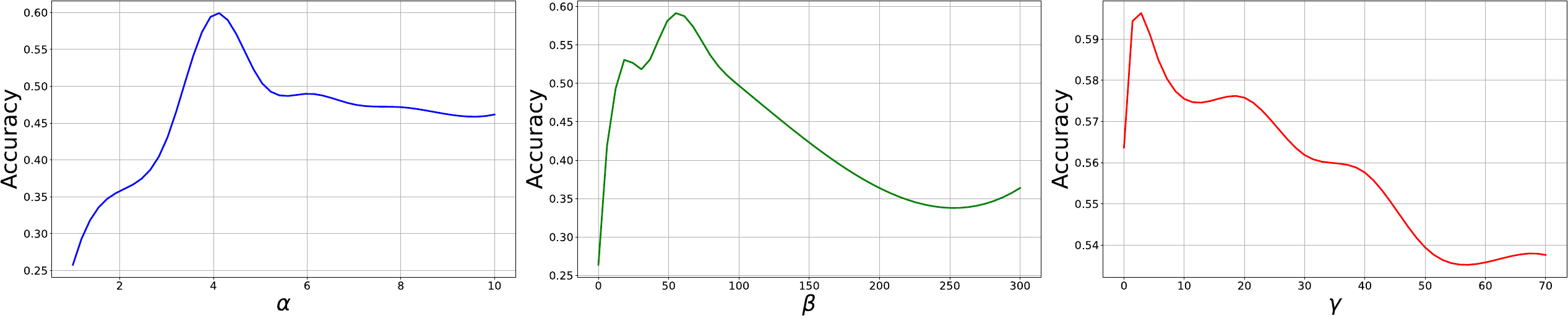}}
\caption{Clustering Accuracy vs. Parameter Variations for Yale Dataset.}
\label{Fig:para}
\end{figure}

We further investigate how the parameters $\alpha$, $\beta$, and $\gamma$ influence clustering performance. The values of $\alpha$, $\beta$, and $\gamma$ are varied within the ranges [1, 10], [0.01, 300], and [0.0001, 70], respectively. Due to the space limitation, only the parameter sensitivity analysis for the Yale dataset is provided in Fig.~\ref{Fig:para}, illustrating the relationship between accuracy and parameter variations.

\subsubsection{Motion Segmentation \label{sec:hop}}
We evaluate DKLM on the Hopkins155 database, a widely used benchmark for motion segmentation. This dataset comprises 155 sequences, each containing 2 or 3 distinct motions (see Fig.~\ref{Fig:realdata}(f)). These sequences are categorized into three groups: 104 indoor checkerboard, 38 outdoor traffic, and 13 articulated/nonrigid sequences. Ground-truth motion labels and feature trajectories (x-, y-coordinates) across frames are provided, ensuring no outliers but incorporating moderate noise. The number of feature trajectories ranges from 39 to 556 per sequence, with frame counts spanning from 15 to 100. 
\begin{figure}[t]
\centerline{\includegraphics[width=0.8\linewidth]{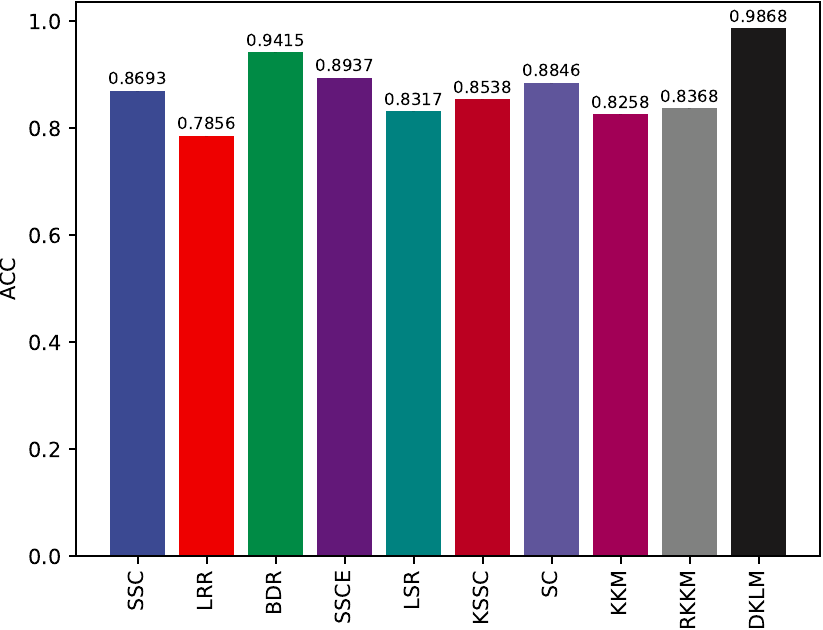}}
\caption{Comparison of Clustering Accuracy on the Hopkins155 Dataset.}
\label{Fig:resulthop}
\end{figure}

Fig.\ref{Fig:resulthop} shows a comparison of clustering accuracy across various methods on Hopkins155, demonstrating that our approach achieves significantly better results. Fig.\ref{Fig:hop} highlights the clustering performance on four specific sequences: people1, cars10, 1R2TCR, and 2T3RTCR. Our method performs effectively on imbalanced clusters (e.g., people1) and handles sequences with noticeable perspective distortion (e.g., 1R2TCR and 2T3RTCR) well. In the latter cases, the affine camera assumption breaks down due to perspective distortion, leading the motion trajectories to lie in nonlinear subspaces.
\begin{figure}[t]
\centerline{\includegraphics[width=1\linewidth]{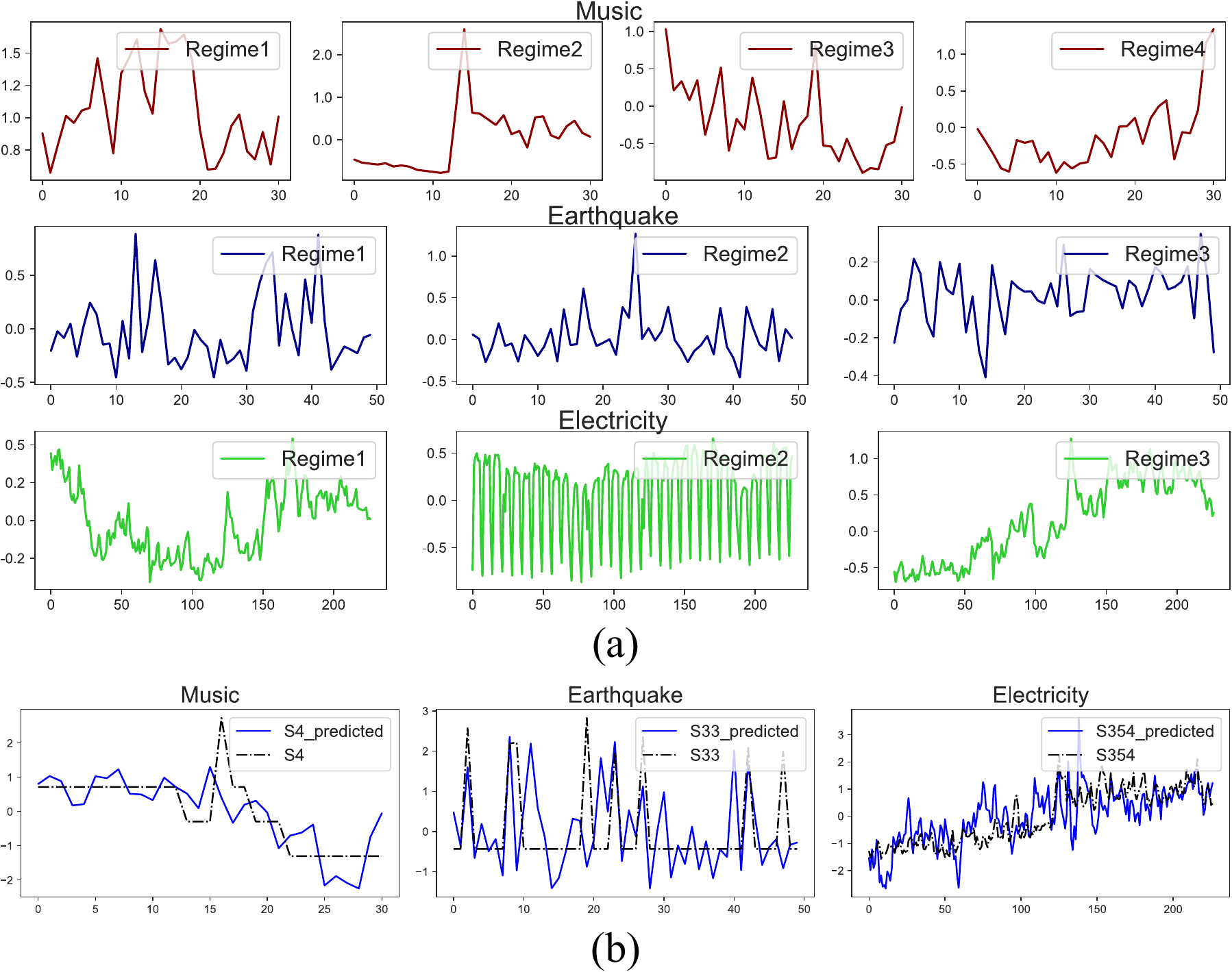}}
\caption{Result of time series. (a) Discovered distinct regimes for the three time series datasets. (b) True (black) and forecasted (blue) values for the three series, each from a real dataset. (c) The best window size (red line) for the three time series datasets.}
\label{Fig:ts}
\end{figure}
\section{DKLM at work \label{sec:ts}}
Here, we demonstrate one of our promising applications: regime identification in time series. Regime identification refers to the segmentation of time series into clusters of similar patterns, which plays a pivotal role in achieving accurate time series forecasts and enhancing the interpretability of the results. We extend DKLM to learn effective representations for capturing changing regimes. Thanks to our proposed method DKLM, each representation accounts for the nonlinear interactions between variables. We perform DKLM on three time-series datasets, \ie Google Music\footnote{\url{http://www.google.com/trends/}}, Earthquake\footnote{\url{https://www.cs.ucr.edu/\%7Eeamonn/time_series_data_2018}} and Electricity\footnote{\url{https://archive.ics.uci.edu/ml/datasets/}}.

Given the $N$ time series, in the learning process, a fixed window slides over all the series and generates $b$ subseries from the beginning of the $N$ time series, each such subseries corresponds to one timestamp of the sliding window. Then, we adopt DKLM to learn a kernel self-representation for subseries in each window and predict the future behaviors representation $\mathbf{Z}^{b+1}$ in the $(b+1)$-$th$ window by regression. Fig.~\ref{Fig:ts}(a) shows the result of regime identification in the three time series. In the Music time series, we identified four distinct regimes, while three regimes were found in both the Earthquake and Electricity datasets. In real-world scenarios, ground truth labels for validating these regimes are often unavailable. However, the accuracy of forecasting, which relies on the identified regimes, provides a practical way to assess their validity. Fig.~\ref{Fig:ts}(b) illustrates the forecasts generated for the three datasets, highlighting the contributions of the discovered regimes in improving forecasting accuracy.


\section{Conclusions}
This paper introduces a novel data-driven kernel subspace clustering framework, enabling the discovery of nonlinear data structures. Unlike conventional approaches that rely on predefined kernels, our method learns the kernel directly from the data’s self-representation while adhering to adaptive-weighting principles and the multiplicative triangle inequality constraint. Additionally, we introduce a negative term in the kernel self-representation process to ensure the preservation of local manifold structures. The proposed model is evaluated through several datasets, and we also extend our study to several real time-series in different applications to demonstrate the practicality and generality of our method. The theoretical and experimental analysis demonstrates the superior performance and robustness of our method.”

\section*{Acknowledgments}
This work was partially supported by the Natural Sciences and Engineering Research Council of Canada (NSERC) under CRD grant CRDPJ 537461-18 and Discovery Accelerator Supplements Grant RGPAS-2020-00089, Quebec Prompt Grant 114\_IA\_Wang-DRC 2019 and the National Natural Science Foundation of China (NSFC) under Grant No.U1805263.

\appendix
\subsection{Permutation Invariance of Representation}\label{proof2}
\begin{proof}[Proof of the Theorem~\ref{P-kernel}]
Given a permutation matrix $\mathbf{P}$, consider the self-representation matrix $\tilde{\mathbf{Z}}$ for the permuted data matrix $\mathbf{X}\mathbf{P}$. The objective for $\mathbf{X}\mathbf{P}$ becomes:
\begin{equation}
\begin{split}
    \min_{\tilde{\mathbf{Z}}} \left\| \Phi(\mathbf{XP}) - \Phi(\mathbf{XP})\tilde{\mathbf{Z}} \right\|^2+\Omega(\tilde{\mathbf{Z}}), \\  \text{s.t.} \ \tilde{\mathbf{Z}} = \tilde{\mathbf{Z}}^\mathrm{T} \geq 0, \ \text{diag}(\tilde{\mathbf{Z}}) = 0
\end{split}
\end{equation}

By the properties of kernel functions and permutation matrices, we have $\Phi(\mathbf{SP})=\Phi(\mathbf{S}P)$. Substituting this into the objective function for $\tilde{\mathbf{Z}}$, we have:
\begin{equation}
\begin{split}
    \min_{\tilde{\mathbf{Z}}} \left\| \Phi(\mathbf{S}\mathbf{P}) - \Phi(\mathbf{X})\mathbf{P}\tilde{\mathbf{Z}} \right\|^2+\Omega(\tilde{\mathbf{Z}}) \\ \text{s.t.} \     \tilde{\mathbf{Z}} = \tilde{\mathbf{Z}}^\mathrm{T} \geq 0, \ \text{diag}(\tilde{\mathbf{Z}}) = 0
    \end{split}
\end{equation}

Since \( P \) is a permutation matrix,$\mathbf{P}\mathbf{P}^\mathrm{T}=\mathbf{I}$, the identity matrix. We apply the transformation $\mathbf{P\tilde{Z}P^\mathrm{T}}$ to the objective function:
\begin{equation}
\begin{split}
    \min_{\tilde{\mathbf{Z}}} \left\| \Phi(\mathbf{X}) - \Phi(\mathbf{X})\mathbf{P\tilde{Z}P^\mathrm{T}} \right\|^2+\Omega(\tilde{\mathbf{Z}}) \\ \text{s.t.} \ \tilde{\mathbf{Z}} = \tilde{\mathbf{Z}}^\mathrm{T} \geq 0, \ \text{diag}(\tilde{\mathbf{Z}}) = 0
 \end{split}
\end{equation}

For the function to be minimized, $\mathbf{P\tilde{Z}P^\mathrm{T}}$ must be the optimal representation matrix for $\mathbf{X}$, which is $\mathbf{Z}$. Therefore, $\mathbf{P\tilde{Z}P^\mathrm{T}} = \mathbf{Z}$, or equivalently, $\tilde{\mathbf{Z}} = \mathbf{P}^\mathrm{T}\mathbf{ZP}$.

This result shows that the self-representation matrix $\mathbf{Z}$ for $\mathbf{X}$ transforms to $ \mathbf{P}^\mathrm{T}\mathbf{ZP}$ for the permuted data matrix $\mathbf{XP}$, demonstrating the invariance of the representation matrix under permutations of the data matrix.

\end{proof}

\bibliographystyle{IEEEtran}
\bibliography{IEEEabrv,mybibliography}

\end{document}